\PassOptionsToPackage{unicode=true}{hyperref} % options for packages loaded elsewhere
\PassOptionsToPackage{hyphens}{url}
\RequirePackage{fix-cm}
\RequirePackage{amsmath}
\documentclass{svjour3} 
\smartqed  % flush right qed marks, e.g. at end of proof
\usepackage{afterpage}
\usepackage{changepage}
\usepackage{booktabs}
\usepackage{threeparttablex}
\usepackage{lscape}
\usepackage{pifont}
\usepackage{threeparttable}
\usepackage{longtable}
\usepackage{lmodern}
\usepackage{amssymb,amsmath}
\usepackage{ifxetex,ifluatex}
\usepackage{fixltx2e}
\usepackage{bigstrut}
\usepackage{tabularx}
\usepackage{graphicx}
\usepackage{algorithm}  
\usepackage{algpseudocode}  
\usepackage{indentfirst}
\usepackage{multicol}
\usepackage{multirow}
\usepackage{url}
\usepackage{booktabs}
\usepackage{color}
\usepackage{cite}
\usepackage{framed}
\usepackage{amssymb}
\usepackage{amsmath}
\usepackage{verbatim}
\usepackage[colorlinks,urlcolor=blue,linkcolor=black,anchorcolor=black,citecolor=black]{hyperref}
\begin{document}

\newcommand{\cmark}{\ding{51}}%
\newcommand{\xmark}{\ding{55}}%

\title{ Machine Reading Comprehension: \\a Literature Review}

\author{Xin Zhang         \and
    An Yang \and Sujian Li \and Yizhong Wang    
}

\institute{Xin Zhang \at
              Peking University\\
              Tel.: +86-10-62753081\\
              \email{zhangxin@pku.edu.cn}           %  \\
           \and
           An Yang \at
               Peking University
           \and
           Sujian Li \at
              Peking University
          \and
           Yizhong Wang \at
             Peking University  
}

\date{Received: date / Accepted: date}

\maketitle

\begin{abstract}
Machine reading comprehension aims to teach machines to understand a text like a human, and is a new challenging direction in Artificial Intelligence.
This article 
%is an overview of the literature on machine reading comprehension, and 
summarizes recent advances in MRC, mainly focusing on  two aspects (i.e., corpus and techniques).
The specific characteristics of various MRC corpus are listed and compared. 
The main idea of some typical MRC techniques are also described. 
%Abstract will be written when the main contents of thesis are finished.
\keywords{Machine Reading Comprehension \and Natural Language Processing \and More}
\end{abstract}

\section{Introduction}
\label{intro}
Over past decades, there has been a growing interest in making the machine understand human languages. And recently, great progress has been made in machine reading comprehension (MRC). 
In one view, the recent tasks titled MRC can also be seen as the extended tasks of question answering (QA). 
% 首先介绍70年代的QA的三种形式，其中最后1种是MRC
\begin{comment}
根据2000年QA的一篇综述：https://www.cambridge.org/core/services/aop-cambridge-core/content/view/95EA883AFC7EB2B8EC050D3920F39DE2/S1351324901002807a.pdf/natural_language_question_answering_the_view_from_here.pdf
对QA的起源进行如下介绍
\end{comment}
\par As early as 1965, Simmons had summarized a dozen of QA systems proposed over the preceding 5 years in his review\cite{simmons1964answering}.
The survey by Hirschman and Gaizauskas\cite{hirschman2001natural} classifies those QA model into three categories, namely the natural language front ends to the database, the dialogue interactive advisory systems and the question answering and story comprehension.
%We follow the classification in another survey\cite{hirschman2001natural} and separate those QA models into three categories, namely the natural language front ends to database, the dialogue interactive advisory systems and the question answering and story comprehension.
For QA systems in the first category, like the BASEBALL\cite{green1961baseball} and the LUNAR\cite{woods1973progress} system, they usually transform the natural language questions into a query against a structured database based on linguistic knowledge. Although performing fairly well on certain tasks, they suffered from the constraints of the narrow domain of the database.
As about the dialogue interactive advisory systems, including the SHRDLU\cite{winograd1972understanding} and the GUS\cite{bobrow1977gus}, early models also used the database as their knowledge source. Problems like ellipsis and anaphora in the conservation, which those systems struggled in dealing with, still remain as a challenge even for nowadays models.
The last category can be seen as the origin of modern MRC tasks.
%Given a story, 
Wendy Lehnert\cite{lehnert1977conceptual} first proposed that the QA systems should consider both the story and the question, and answer the question after necessary interpretation and inference. Lehnert also designed a system called QUALM\cite{lehnert1977conceptual} according to her theory.

% 再介绍MRC在近年来的发展，也即简要概括本文主体中的corpus，techniques即可
\par The past decade has witnessed a huge development in the MRC field, including the soar of numbers of corpus and great progress in techniques. 
\par As about MRC corpus, plenty of datasets in different domains and styles have been released in recent years. In 2013, MCTest\cite{richardson2013mctest} was released as a multiple-choice reading comprehension dataset, which was of high quality whereas too small to train  neural models. In 2015, CNN/Daily Mail\cite{ref_cnn} and CBT\cite{ref_cbt} were released. These two datasets were generated automatically from 
different%according 
domains and much larger than previous datasets. In 2016, SQuAD\cite{ref_squad} was shown up as the first large-scale dataset with questions and answers written by the human. Many techniques have been proposed along with the competition on this dataset. In the same year, the MS MARCO\cite{nguyen2016ms} was released with the emphasis on narrative answers. Subsequently, NewsQA\cite{ref_newsqa} and NarrativeQA\cite{kovcisky2018narrativeqa} were constructed in similar paradigm with SQuAD and MS MARCO respectively. And both datasets were crowdsourced with the expectation for high quality. Next, various datasets sourced from different domains sprung up in the following two years, including RACE\cite{lai2017large}, CLOTH\cite{xie2017cloth} and ARC\cite{clark2018arc} that were collected from exams, TriviaQA\cite{ref_triviaqa} that were based on trivias and MCScript\cite{ostermann2018mcscript} primarily focused on scripts. Released in 2018, WikiHop\cite{ref_wikihop} aimed at examing systems' ability of multi-hop reasoning, and CoQA\cite{reddy2018coqa} were proposed to test conversation ability of models.
\par The appearance of large-scale datasets above makes training an end-to-end neural MRC model possible. When competing on the leaderboard, many models and techniques were developed in an attempt to conquer a certain dataset. From word representations, attention mechanisms to high-level architectures, neural models evolve rapidly and even surpass human performance in some tasks.

In this article, we aim to make an extensive review on recent datasets and techniques for MRC. 
In Section \ref{sec:corpus}, we categorize the MRC datasets into three types and describe them briefly.
In Section \ref{sec:3}, we introduce the traditional non-neural methods, neural network based models and attention mechanism which have been used in the MRC tasks.
Finally, Section 4 concludes our review.

% 任务分类
% *** This section can be removed ? ***
%% already removed
% \section{Classification of tasks in MRC}
% \label{sec:1}
% Generally speaking, there are three major types of tasks in the area of Machine Reading Comprehension.

% 数据集
\section{MRC Corpus}
\label{sec:corpus}
% 简要的引入
%To accelerate progresses in the certain field, large and realistic datasets have been developed in many areas. 
The fast development of the MRC field is driven by various large and realistic datasets released in recent years.
Each dataset is usually composed of documents and questions for testing the document understanding ability. The answers for the raised questions can be obtained through seeking from the documents or selecting the preseted options.
%According to the formats of answers, we classify the datasets into three types, namely extraction answer, description answer and multiple-choice answer, and introduce them respectively as follows.
Here, 
according to the formats of answers, we classify the datasets into three types, namely datasets with extractive answers, with descriptive answers and with multiple-choice answers, and introduce them respectively in the following subsections.In parallel to this survey, there are also new datasets \cite{hotpotqa, drop, googlenaturalquestions} steadily coming out with more diverse task formulations, and testing more complicated understanding and reasoning abilities. 

%目前出现了很多数据集，这些数据集可以从不同维度进行分类介绍，例如针对不同的MRC任务、问题的答案形式、。。。这里我们根据问题答案的表现形式进行分类，大体分成了 text span形式、描述形式和多选择形式 三个类别去介绍。

%Famous examples include ImageNet for object-recognition \cite{ref_imagenet} and the Penn Treebank for syntactic parsing\cite{ref_treebank}. Similarly, 
%Many datasets of different genres have been  released for MRC task recently and are introduced detailedly as follows. 
%%%%%%%%%%%%%%%%%%%%%%%%%%%
%%% 抽取式的数据集的罗列
%%%%%%%%%%%%%%%%%%%%%%%%%%%
\subsection{Datasets With Extractive Answers}
To test a system's ability of reading comprehension, this kind of datasets, which originates from Cloze\cite{ref_cloze} style questions, firstly provide the system with a large amount of documents or passages, and then feed it with questions whose answers are segments of corresponding passages. A good system should select a correct text span from a given context. Such comprehension tests are appealing because they are objectively gradable and may measure a range of important abilities, from basic understanding to complex inference \cite{ref_richardson}.
%***add some words to generalize the following datasets***
% added as follows
\par Either sourced by crowdworkers or generated automatically from different corpus, these datasets all use a text span in the document as the answer to the proposed question. Many of them released in recent years are large enough for training strong neural models. These datasets include SQuAD, CNN/Daily Mail, CBT, NewsQA, TriviaQA, WIKIHOP which are described briefly below.

%%%%%%%%% SQuAD
\paragraph{SQuAD} 
One of the most famous dataset of this kind is Stanford Question Answering Dataset (SQuAD)~\cite{ref_squad}. The Stanford Question Answering Dataset v1.0 (SQuAD v1.0) \footnote{https://stanford-qa.com} consists of questions posed by crowdworkers on a set of Wikipedia articles, where the answer to each question is a segment of text (or span) from the corresponding reading passage. SQuAD v1.0 contains 107,785 question-answer pairs from 536 articles, which is much larger than previous manually labeled RC datasets. We quote some example question-answer pairs as in Fig.\ref{fig:example}, where each answer is a span of the document.

\begin{figure}[ht]
  \fontfamily{cmss}\selectfont
  \footnotesize
  \centering
  \rule{\linewidth}{1pt}
  \begin{quotation}
    In meteorology, precipitation is any product of the condensation of atmospheric water vapor that falls under {\color{red} gravity}. The main forms of precipitation include drizzle, rain, sleet, snow, {\color{green} graupel} and hail... Precipitation forms as smaller droplets coalesce via collision with other rain drops or ice crystals {\color{blue} within a cloud}. Short, intense periods of rain in scattered locations are called ``showers''.
  \end{quotation}
    \vspace{2mm}
    \begin{tabular}{p{10.3cm}}
    What causes precipitation to fall? \\
    {\color{red} gravity} \\
    \vspace{0.4mm}
    What is another main form of precipitation besides drizzle, rain, snow, sleet and hail? \\
    {\color{green} graupel} \\
    \vspace{0.4mm}
    Where do water droplets collide with ice crystals to form precipitation? \\
    {\color{blue} within a cloud}
    \vspace{2mm}
    \end{tabular}
  \rule{\linewidth}{1pt}
  \caption{
     Question-answer pairs for a sample passage in the SQuAD\cite{ref_squad}.
    }
  \label{fig:example}
\end{figure}

In SQuAD v1.0 \cite{ref_squad}, the answers belong to different categories as shown in Table \ref{tab:AnsPosCategory}. As we can see, common noun phrases make up 31.8\% of the whole data, proper noun phrases \footnote{consisting of person, location and other entities} make up 32.6\% of the data, and the rest one third consists of date, numbers, adjective phrase, verb phrase, clauses and so on. This indicates that the answers of SQuAD v1.0 displays reasonable diversity.
As about the reasoning skills of SQuAD v1.0 to answer the questions, the authors show that all examples at least have some lexical or syntactic divergence between the question and the answer in the passage, through manually annotating some examples.

\begin{table}
  \centering
  \setlength{\tabcolsep}{12mm}{
  \begin{tabular}{l@{\hskip 0.5em} r l}
    \toprule
       Answer type  & Percentage & Example \\
    \midrule
       Date & 8.9\% & 19 October 1512 \\
       Other Numeric & 10.9\% & 12 \\
       Person & 12.9\%  & Thomas Coke \\
       Location & 4.4\% &  Germany \\
       Other Entity & 15.3\% &  ABC Sports \\
       Common Noun Phrase & 31.8\% & property damage \\
       Adjective Phrase & 3.9\% & second-largest \\
       Verb Phrase & 5.5\% & returned to Earth \\
       Clause & 3.7\% & to avoid trivialization \\
       Other & 2.7\% & quietly \\
    \bottomrule
  \end{tabular}}
  \caption{
    Answer type distribution in SQuAD\cite{ref_squad} 
    }
  \label{tab:AnsPosCategory}
\end{table}

%also gives some analysis of answer types as in Fig.\ref{squad_pic2}, which indicates that answers in the dataset display reasonable diversity.
%% 已经添加
%\par When it comes to the reasoning skills required to answer the questions, some randomly sampled questions are manually annotated. The result given by the paper shows that all examples at least have some lexical or syntactic divergence between the question and the answer in the passage.

% 图1到图4 需要重新画图或者把图去了，把一些总结的话放到论文中。
%% 部分留下来的图已经重新作图，其他的换成总结的话了
Later, SQuAD v2.0\cite{ref_squad_2} was released with emphasis on unanswerable questions. This new version of SQuAD adds over 50,000 unanswerable questions which were created adversarially by crowdworkers according to the original ones. In order to challenge the existing models which tend to make unreliable guesses on questions whose answers are not stated in context, newly added questions are highly similar to corresponding context and have plausible (but incorrect) answers in context. 
We also quote some examples as shown in Fig.\ref{squad_unanswerable}.
The unanswerable questions in SQuAD v2.0 are posed by humans, and exhibit much more diversity and fidelity than those in other automatic constructed datasets  \cite{ref_addsent,ref_zero_shot}.
In such cases, simple heuristics 
%which are based on over
%These features prevent simple heuristics, 
which are based on overlapping\cite{ref_overlap} or entity type recognition\cite{ref_type_recog}, are not able to distinguish answerable from unanswerable questions.

\begin{figure}[t]
  \begin{framed}
  \footnotesize
    \textbf{Article:} Endangered Species Act

    \textbf{Paragraph:}
    ``{
      \dots Other legislation followed, including the Migratory Bird Conservation Act of 1929, a \textcolor{blue}{1937 treaty} prohibiting the hunting of right and gray whales, and the \textcolor{red}{Bald Eagle Protection Act of 1940}. These \textcolor{red}{later laws} had a low cost to society---the species were relatively rare---and little \textcolor{blue}{opposition} was raised.}''

    \vspace{0.08in}

    \textbf{Question 1:} ``Which laws
    faced significant \textcolor{blue}{opposition}?''

    \textbf{Plausible Answer:} \textit{\textcolor{red}{later laws}}
    
    \vspace{0.08in}

    \textbf{Question 2:} ``What was the name of the \textcolor{blue}{1937 treaty}?''

    \textbf{Plausible Answer:} \textit{\textcolor{red}{Bald Eagle Protection Act}}

  \end{framed}
  \caption{ Unanswerable question examples with plausible (but incorrect) answers\cite{ref_squad_2}
  }
  \label{squad_unanswerable}
\end{figure}

\paragraph{CNN/Daily Mail}
CNN and Daily Mail Dataset\cite{ref_cnn}, which was released by Google DeepMind and University of Oxford in 2015, is the first large-scale reading comprehension dataset constructed from  natural language materials. Unlike  most relevant work which uses templates or syntactic/semantic rules to extract document-query-answer triples, this work collects 93k articles from the CNN\footnote{www.cnn.com} and 220k articles from the Daily Mail\footnote{www.dailymail.co.uk}  as the source text. Since each article comes along with a number of bullet points to summarize the article, this work converts these bullet points into document-query-answer triples with the Cloze\cite{ref_cloze} style questions.

To exclusively examine a system's ability of reading comprehension rather than using world knowledge or co-occurrence, further modifications are implemented on those triples to construct an anonymized version. That is, each entity is anonymized by using an abstract entity marker, which is not easily predicted by using world knowledge or n-gram language model.
%Firstly, each entity is anonymized by using an abstract entity marker, so under this circumstance one can not predict answers by only using requisite world knowledge. Secondly, whenever a data point is loaded, those entity markers will permute randomly, so n-gram language model can not  perform well by simply learning some word co-occurrences. 
An example data point and its anonymized version is shown in Table \ref{cnn_datapoint}.

Some basic corpus statistics of CNN and Daily Mail are shown in Table \ref{tab:corpora}. 
We also quote the percentages of the right answers appearing in the top N most frequent entities in an given document as in Table \ref{tab:cloze_hardness}, illustrating the difficulty degree of the questions to some extent.
%The character presented in Table \ref{tab:cloze_hardness}, namely the percentages of the right answers appearing in the top N most frequent entities in an given document, illustrates the difficulty of those questions to some extent.

% cnn的sample
\begin{table}
\centering
\begin{footnotesize}
\begin{tabular}{p{0.45\textwidth}p{0.45\textwidth}}
\toprule
\textbf{Original Version} & \textbf{Anonymised Version} 
 \\ \hline
\textbf{Context} \\ 
The BBC producer allegedly struck by Jeremy Clarkson will not press charges against the ``Top Gear'' host, his lawyer said Friday. Clarkson, who hosted one of the most-watched television shows in the world, was dropped by the BBC Wednesday after an internal investigation by the British broadcaster found he had subjected producer Oisin Tymon ``to an unprovoked physical and verbal attack.'' \dots
& 
the \textit{ent381} producer allegedly struck by \textit{ent212} will not press charges against the `` \textit{ent153} '' host , his lawyer said friday .  \textit{ent212} , who hosted one of the most - watched television shows in the world , was dropped by the \textit{ent381} wednesday after an internal investigation by the \textit{ent180} broadcaster found he had subjected producer \textit{ent193} `` to an unprovoked physical and verbal attack . '' \dots 
\\ \midrule
\textbf{Query} \\ 
Producer \textbf{X} will not press charges against Jeremy Clarkson, his lawyer says. & producer \textbf{X} will not press charges against \textit{ent212} , his lawyer says.
\\ \midrule
\textbf{Answer} \\
Oisin Tymon  & \textit{ent193} 
\\ \bottomrule
\end{tabular}
\end{footnotesize}
\caption{An example data point quoted from \cite{ref_cnn}}
\label{cnn_datapoint}
\end{table}

\begin{table}[t]
\footnotesize
{\centering
  \begin{minipage}[t]{0.67\textwidth}
  \centering
  \begin{tabular}[t]{@{}l@{~}r@{~~}r@{~~}r@{}l@{}r@{~~}r@{~~}r@{}}
    \toprule
    & \multicolumn{3}{c}{{\bf CNN}} &\phantom{aa}& \multicolumn{3}{c}{{\bf
Daily Mail}} \\
    \cmidrule{2-4} \cmidrule{6-8}
    & train & valid & test && train & valid & test \\
    \midrule
    \# months    & 95       & 1     & 1     &&      56 & 1      & 1 \\
    \# documents &  90,266  & 1,220 & 1,093 && 196,961 & 12,148 & 10,397 \\
    \# queries   & 380,298  & 3,924 & 3,198 && 879,450 & 64,835 & 53,182 \\
    Max \# entities & 527   & 187   & 396   && 371     & 232    & 245 \\
    Avg \# entities & 26.4  & 26.5  & 24.5  && 26.5    & 25.5   & 26.0 \\
    Avg \# tokens  & 762    & 763   & 716   && 813     & 774    & 780  \\
    Vocab size & \multicolumn{3}{c}{{118,497}} && \multicolumn{3}{c}{{208,045}} \\
    \bottomrule
  \end{tabular}
  \caption{Corpus statistics of CNN and Daily Mail \cite{ref_cnn}}
  \label{tab:corpora}
\end{minipage}
\hfill
\begin{minipage}[t]{0.31\textwidth}
\footnotesize
  \centering
  \begin{tabular}[t]{@{}l@{~~}r@{~~}r@{}}
    \toprule
    \textbf{Top N} & \multicolumn{2}{c}{{\bf Cumulative \%}} \\
    \cmidrule{2-3}
    & \textbf{CNN} & \textbf{Daily Mail} \\
    \midrule
    1  & 30.5 & 25.6 \\
    2  & 47.7 & 42.4 \\
    3  & 58.1 & 53.7 \\
    5  & 70.6 & 68.1 \\
    10 & 85.1 & 85.5 \\
    \bottomrule
  \end{tabular}
  \caption{Percentage of correct Answers contained in the top $N$ most frequent entities in a given document quoted from \cite{ref_cnn}.}
  \label{tab:cloze_hardness}
\end{minipage}
}
\end{table}

% cnn的图片
% \begin{figure}
% \centering
% \includegraphics[width=\textwidth]{cnn_pic_1.png}
% \caption{an example of a data point presented in paper\cite{ref_cnn} }
% \label{cnn_pic_1}
% \end{figure}
% \begin{figure}
% \centering
% \includegraphics[width=\textwidth]{cnn_pic_2.png}
% \caption{ basic statistics presented in paper\cite{ref_cnn} }
% \label{cnn_pic_2}
% \end{figure}
% \begin{figure}
% \centering
% \includegraphics[width=0.3\textwidth]{cnn_pic_3.png}
% \caption{}
% \label{cnn_pic_3}
% \end{figure}
% 都换成表格了

\begin{figure}
\centering
\includegraphics[width=\textwidth]{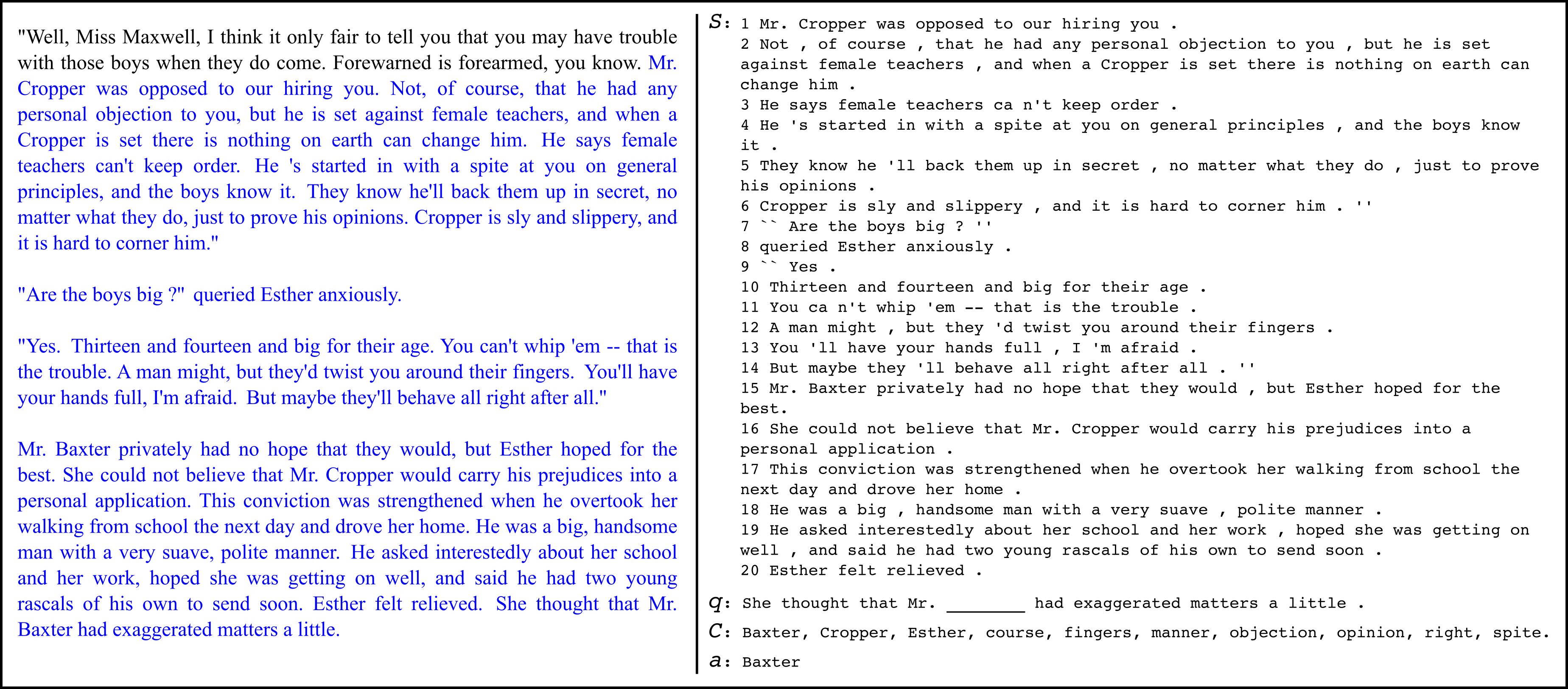}
\caption{An CBT example quoted from \cite{ref_cbt}}
\label{cbt_pic_1}
\end{figure}

%%%%%%%%%
\paragraph{CBT} The Children's Book Test\cite{ref_cbt} is a part of \textit{bAbI} project of Facebook AI Research\footnote{https://research.fb.com/downloads/babi/} which aims at researching automatic text understanding and reasoning. Children books are chosen because they ensure a clear narrative structure which aids this task. 
The children stories used in CBT  come from books freely available from \textit{Project Guntenberg}\footnote{https://www.gutenberg.org}. 
Questions are formed by enumerating 21 consecutive sentences from chapters in books, of which the first 20 sentences serve as \textit{context}, and the last one as \textit{query} after removing one word. 10 candidates are selected from words appearing in either context or query. An example question is given in Fig. \ref{cbt_pic_1} and the dataset size is shown in Table \ref{tab:cbt_stat}.

%It distinguishes itself from other tasks of predicting syntactic function words by predicting lower-frequency words in certain context, and requires more semantic content to be captured. In other words, it requires a better understanding of a wider linguistic context. 
%This also leads to a finer-grained analysis of this dataset. 
In CBT, four distinct types of word: Named Entities, (Common) Nouns, Verbs and Prepositions\footnote{based on output from the POS tagger and named-entity-recognizer in the Stanford Core NLP Toolkit\cite{ref_SCNLP}.}, are removed respectively to form 4 classes of questions. For each class of questions, the nine wrong candidates are selected randomly from words which have the same type as the answer options in the corresponding context and query. 
%The exact size of this dataset is shown in Fig.\ref{cbt_pic_2}.
%This also lead to a finer-grained analysis of this dataset. Four distinct types of word: Named Entities, (Common) Nouns, Verbs and Prepositions\footnote{based on output from the POS tagger and named-entity-recogniser in the Stanford Core NLP Toolkit\cite{ref_SCNLP}.} are removed respectively and thus form 4 classes of questions. For each class of questions, the nine wrong candidates are selected randomly from words which have the same type as the answer in the corresponding context and query. The exact size of this dataset is shown in Fig.\ref{cbt_pic_2}.

%One interesting phenomenon the authors observed is that, 
Compared to human performance on this dataset, the state-of-art models like Recurrent Neural Networks (RNNs) with Long-Short Term Memory (LSTM)\cite{hochreiter1997long} performed much worse  when predicting nouns or named entities, whereas they did great job in predicting prepostions and verbs. This may probably be explained by the fact that these models are almost based exclusively on local contexts. In contrast, Memory Networks\cite{ref_mem_net}
%, the \textit{contextual models}, 
can exploit a wider context  and outperform the conventional models when predicting nouns or named entities.
Thus, this corpus encourages the use of world knowledge in comparison with \textit{CNN/Daily Mail}, and therefore focuses less on paraphrasing parts of a context.

% cbt的所有图片
% 上面这个保留，但是替换了原图

%
% \begin{figure}
% \centering
% \includegraphics[width=\textwidth]{cbt_pic_2.png}
% \caption{statistics of CBT given in paper\cite{ref_cbt}}
% \label{cbt_pic_2}
% \end{figure}
% stat图换成了表格
\vspace{-2mm}
\begin{table}[ht]
  \begin{center}
    {\small 
      {\sc 
        \begin{tabular}{l|ccc}
        \hline
          & Training & Validation & Test \\
          \hline
          \hline
          Number of books & 98 & 5 & 5 \\
          Number of questions (context+query)& 669,343 & 8,000 & 10,000  \\
          Average words in contexts & 465 & 435 & 445 \\
          Average words in queries & 31 & 27 & 29 \\
          Distinct candidates & 37,242 & 5,485 & 7,108 \\
          \hline
          Vocabulary size & \multicolumn{3}{|c}{53,628}\\
          \hline
        \end{tabular}
      }
    }
    \caption{\label{tab:cbt_stat}  Corpus statistics of CBT \cite{ref_cbt}}
  \end{center}
\vspace*{-4ex}
\end{table}

%%%%%%%%%
\paragraph{NewsQA} Based on 12,744 news articles from CNN\footnote{www.cnn.com} news, the \textit{NewsQA}\cite{ref_newsqa} dataset contains 119,633 question-answer pairs generated by crowdworkers. Similar to SQuAD\cite{ref_squad}, the answer to each question is a text span of arbitrary length in the corresponding article (a \textit{null} span is also included). 
CNN articles are chosen as source materials, because in the authors' view, machine comprehension systems are particularly suited to high-volume, rapidly changing information sources like news\cite{ref_newsqa}. 
The major differences between CNN/Daily Mail and NewsQA are that, the answers of NewsQA are not necessarily entities and therefore no anonymization procedure is considered in the generation of NewsQA. 

%The authors asserted that NewsQA is more challenging to current models than SQuAD, and thus can further push the development of MRC systems.

The statistics of answer types in NewsQA is shown in Table \ref{tab:news-answer-type}. As can be seen in the table, the variety of answer types is ensured. Furthermore, the authors sampled 1000 examples from NewsQA and SQuAD respectively and analyzed the possible reasoning skills to answer the questions. The results indicate that compared to SQuAD, a larger proportion of questions in NewsQA require high-level reasoning skills, including \textit{Inference} and \textit{Synthesis}. Besides, %while simple skills like \textit{word matching} and \textit{paraphrasing} make up  most part of both datasets, they constitute less proportion of NewsQA than SQuAD. The detailed comparison result is given in Table \ref{tab:compare_newsqa_squad}. 
while simple skills like \textit{word matching} and \textit{paraphrasing} can solve  most questions in both datasets, NewsQA tends to require more complex reasoning skills than SQuAD.  The detailed comparison result is given in Table \ref{tab:compare_newsqa_squad}.

% newsqa的所有图片
% \begin{figure}
% \centering
% \includegraphics[width=\textwidth]{news_pic_1.png}
% \caption{answer types statistics given in paper\cite{ref_newsqa}}
% \label{news_pic_1}
% \end{figure}
% 上面的答案类型分布的图换成如下的表格
\begin{table}
	\scriptsize
	\centering
    \vspace{4pt}
    \begin{tabular}{lp{0.5\textwidth}cp{0.4\textwidth}cp{0.3\textwidth}}
    	\toprule
    	Answer type & Example & Proportion (\%) \\ \midrule
    	Date/Time & March 12, 2008 & 2.9 \\
		  Numeric & 24.3 million & 9.8 \\
		  Person & Ludwig van Beethoven & 14.8 \\
		  Location & Torrance, California & 7.8 \\
		  Other Entity & Pew Hispanic Center & 5.8 \\
		  Common Noun Phr. & federal prosecutors & 22.2 \\
		  Adjective Phr. & 5-hour & 1.9 \\
		  Verb Phr. & suffered minor damage & 1.4 \\
		  Clause Phr. & trampling on human rights & 18.3 \\
		  Prepositional Phr. & in the attack & 3.8 \\
		  Other & nearly half & 11.2 \\
    	\bottomrule
    \end{tabular}
	\label{tab:news-answer-type}
	\caption{Answer types distribution of \emph{NewsQA}\cite{ref_newsqa}}
\end{table}

% \begin{figure}
% \centering
% \includegraphics[width=\textwidth]{news_pic_3.png}
% \caption{reasoning skill types are a variation on the taxonomy presented by Chen et al. (2016)\cite{ref_exm_cnn} in their analysis of the CNN/Daily Mail dataset. Types are as listed in ascending order of difficulty.}
% \label{news_pic_3}
% \end{figure}
% 上面的reasoning-skills的图换成了一段文字叙述+如下的原始的表格
\begin{table*}[t]
	\scriptsize
	\centering
    \vspace{4pt}
    \begin{tabularx}{\textwidth}{ l X c c}
    	\toprule
    	{Reasoning} & {Example} & \multicolumn{2}{c}{Proportion (\%)} \\
      & & \emph{NewsQA} & \emph{SQuAD} \\
    	\midrule
    	Word Matching &
    	Q: {\bf When were} the {\bf findings} {\bf published}? \newline
    	S: Both sets of research {\bf findings} {\bf were} {\bf published Thursday}... 
    	& 32.7 & 39.8 \\
    	\midrule
		Paraphrasing &
		Q: {\bf Who} is the {\bf struggle between} in Rwanda? \newline
		S: The {\bf struggle} {\bf pits} {\bf ethnic Tutsis}, supported by Rwanda, {\bf against ethnic Hutu}, backed by Congo.
		& 27.0 & 34.3 \\
		\midrule
		Inference &
		Q: {\bf Who} drew {\bf inspiration} from {\bf presidents}? \newline
		S: {\bf Rudy Ruiz} says the lives of US {\bf presidents} can make them {\bf positive role models} for students.
		&  13.2 & 8.6 \\
		\midrule
		Synthesis &
		Q: {\bf Where} is {\bf Brittanee Drexel} from? \newline
		S: The mother of a 17-year-old {\bf Rochester}, {\bf New York} high school student ... says she did not give her daughter permission to go on the trip. {\bf Brittanee} Marie {\bf Drexel}'s mom says...
		& 20.7 & 11.9 \\
		\midrule
		Ambiguous/Insufficient &
		Q: {\bf Whose mother} is {\bf moving} to the White House? \newline
		S: ... {\bf Barack Obama's mother-in-law}, Marian Robinson, will {\bf join} the Obamas at the {\bf family's private quarters} at 1600 Pennsylvania Avenue. [Michelle is never mentioned]
		& 6.4 & 5.4 \\
    	\bottomrule
    \end{tabularx}
    \caption{Reasoning skills used in NewsQA and SQuAD and their corresponding proportions \cite{ref_newsqa}}
	\label{tab:compare_newsqa_squad}
\end{table*}

%%%%%%%%%
\paragraph{TriviaQA} Instead of relying on crowdworkers to create question-answer pairs from selected passages like NewsQA and SQuAD, over 650K TriviaQA\cite{ref_triviaqa} question-answer-evidence triples are generated through automatic procedures. Firstly, a huge amount of question-answer pairs from 14 trivia and quiz-league websites are gathered and filtered. Then the evidence documents for each question-answer pair are collected from either web search results or Wikipedia articles. % of entities.
%in the question independently. 
%The authors assumes that the presence of the answer string in an evidence document means that this document \textit{does} answer the question, which is correct over 75\% of the time\cite{ref_triviaqa}. 
%With the observation that the presence of the answer string in an evidence document means that this document can answer the question  over 75\% of the time\cite{ref_triviaqa}. 
Finally, a clean, noise-free and human-annotated subset of 1975 triples from TriviaQA is given and an triple example is shown in Fig. \ref{trivia_pic_1}.

The basic statistics of TriviaQA is given in Table \ref{tab:triviaqa_statistics}. By sampling 200 examples from the dataset and annotating them manually, 
it turns out that the \textit{Wikipedia titles} (including person, organization, location, and miscellaneous) consists of over 90\% of all answer, and the rest small percentage of answers mainly belong to \textit{Numerical} and \textit{Free Text} type.
The average number of entities per question and the percentages of certain types of questions are also shown in Table \ref{tab:trivia_quest_analysis}.
%it turns out that \textit{Wikipedia Title} (including person, organization, location, and miscellaneous) makes up over 90\% of all answer, and the rest small part consists of \textit{Numerical} and \textit{Free Text} type answers. 
%The hierarchical WordNet synsets of the answer entities are shown in Fig.\ref{trivia_pic_4}, showing that answers have reasonable diversity. Some extra properties like average number of entities per question and the percentages of certain types of questions are also shown in Table \ref{tab:trivia_quest_analysis}.

% triviaqa的所有图片
% \begin{figure}
% \centering
% \includegraphics[width=0.6\textwidth]{trivia_pic_1.png}
% \caption{an triple example given in paper\cite{ref_triviaqa}}
% \label{trivia_pic_1}
% \end{figure}
% 上面这个example的截图换成了如下表格
\begin{figure}[t]
\small
\begin{tabular}{p{11cm}}
\hline
\textbf{Question}: The Dodecanese Campaign of WWII that was an attempt by the Allied forces to capture islands in the Aegean Sea was the inspiration for which acclaimed 1961 commando film? \\
\textbf{Answer}: The Guns of Navarone\\
\textbf{Excerpt}: The Dodecanese Campaign of World War II was an attempt by Allied forces to capture the Italian-held Dodecanese islands in the Aegean Sea following the surrender of Italy in September 1943, and use them as bases against the German-controlled Balkans. The failed campaign, and in particular the Battle of Leros, inspired the 1957 novel \textbf{The Guns of Navarone} and the successful 1961 movie of the same name. \\
\\
\textbf{Question}: American Callan Pinckney's eponymously named system became a best-selling (1980s-2000s) book/video franchise in what genre? \\
\textbf{Answer}: Fitness\\
\textbf{Excerpt}: Callan Pinckney was an American fitness professional. She achieved unprecedented success with her Callanetics exercises. Her 9 books all became international best-sellers and the video series that followed went on to sell over 6 million copies. Pinckney's first video release "Callanetics: 10 Years Younger In 10 Hours" outsold every other \textbf{fitness} video in the US. \vspace{1pt}\\
\hline
\end{tabular}
    \caption{Example question-answer-evidence triples in TriviaQA quoted from\cite{ref_triviaqa}}
    \label{trivia_pic_1}
\end{figure}

% \begin{figure}
% \centering
% \includegraphics[width=0.6\textwidth]{trivia_pic_2.png}
% \caption{basic statistics given in paper\cite{ref_triviaqa}}
% \label{trivia_pic_2}
% \end{figure}
% triviaqa的基本statistic，换成如下的表格了
\begin{table}
\begin{center}
\begin{tabular}{p{0.5\textwidth}p{0.4\textwidth}}
\toprule
Total number of QA pairs & 95,956\\ 
Number of unique answers & 40,478\\
Number of evidence documents & 662,659 \\\midrule
Avg. question length (word) & 14\\ 
Avg. document length (word) & 2,895\\ 
\toprule
\end{tabular}
\end{center}
\caption{Corpus statistics of TriviaQA\cite{ref_triviaqa}. }
\label{tab:triviaqa_statistics}
\end{table}

% \begin{figure}
% \centering
% \includegraphics[width=0.6\textwidth]{trivia_pic_3.png}
% \caption{distribution of answer types given in paper\cite{ref_triviaqa}}
% \label{trivia_pic_3}
% \end{figure}
% 上面这个答案的分布图换为语言描述了

%\begin{figure}
%\centering
%\includegraphics[width=0.6\textwidth]{trivia_pic_4.png}
%\caption{Distribution of Wordnet synsets for entities in the answers\cite{ref_triviaqa}}
%\label{trivia_pic_4}
%\end{figure}
% 上面这个是wordnet图，保留了原始大图，是否需要是个问题...

% \begin{figure}
% \centering
% \includegraphics[width=\textwidth]{trivia_pic_5.png}
% \caption{property annotation result given in paper\cite{ref_triviaqa}}
% \label{trivia_pic_5}
% \end{figure}
% triviaqa额外的一些property，换成如下的原始的表格了
\begin{table*}
\begin{center}
\tiny
\begin{tabular}{l@{}l@{}l}
\toprule
Property & Example annotation & Statistics \\\midrule
Avg. entities/question & Which politician won the \textbf{Nobel Peace Prize} in 2009? & 1.77 per question \\
Fine grained answer type & What \textbf{fragrant essential oil} is obtained from  Damask Rose? & 73.5\% of questions\\
Coarse grained answer type & \textbf{Who} won the Nobel Peace Prize in 2009? & 15.5\% of questions\\
Time frame & What was photographed for the first time in \textbf{October 1959} & 34\% of questions\\
Comparisons & What is the appropriate name of the \textbf{largest} type of frog? & 9\% of questions \\\toprule
\end{tabular}
\end{center}
%\caption{Properties of questions on 200 annotated examples show that a majority of TriviaQA questions contain multiple entities. The boldfaced words hint at the presence of corresponding property.}
\caption{Properties of questions on 200 sampled examples. The boldfaced words mean the presence of the corresponding properties.}
\label{tab:trivia_quest_analysis}
\end{table*}

%
% 这儿可以加一个总结，对这儿总结的数据集从各个属性上列一个表格比较一下
% 我想到的属性：
% size(#passage #questions Avarge )， domain， automatic/manual,  answer type, 
% 以上数据集，目前的修改没有完全去了方法的结论，后面看情况决定是否只保留数据介绍的内容。

%%wikihop 给出了candidate，是一种选择题的变形
\paragraph{WIKIHOP} 
%For the purpose of evaluating a system's ability of multi-hop reasoning across multiple documents, WIKIHOP\cite{ref_wikihop} was released as the first large-scale dataset in this specific area in 2018. 
WIKIHOP\cite{ref_wikihop} was released For the purpose of evaluating a system's ability of multi-hop reasoning across multiple documents in 2018. 
In most existing datasets, the information needed to answer a question is usually contained  in only one sentence, which makes current MRC models pay much attention on simple reasoning skills like locating, matching or aligning information between query and support text. For example, in SQuAD, the sentence which has the highest lexical similarity with the question contains the answer about 80\% of the time\cite{ref_wad}, and a simple binary word-in-query indicator feature boosted the relative accuracy of a baseline model by 27.9\%\cite{ref_weis}.
To move beyond this, the authors define a novel MRC task in which a model needs to combine evidences in different documents to answer the questions. A sample in WIKIHOP which displays such characteristics is shown in Fig.\ref{wikihop_sample}.

\par
\begin{figure}[t]
    \centering
    \includegraphics[width=0.6\columnwidth,trim={0 0.1cm 0 0.1cm},clip]{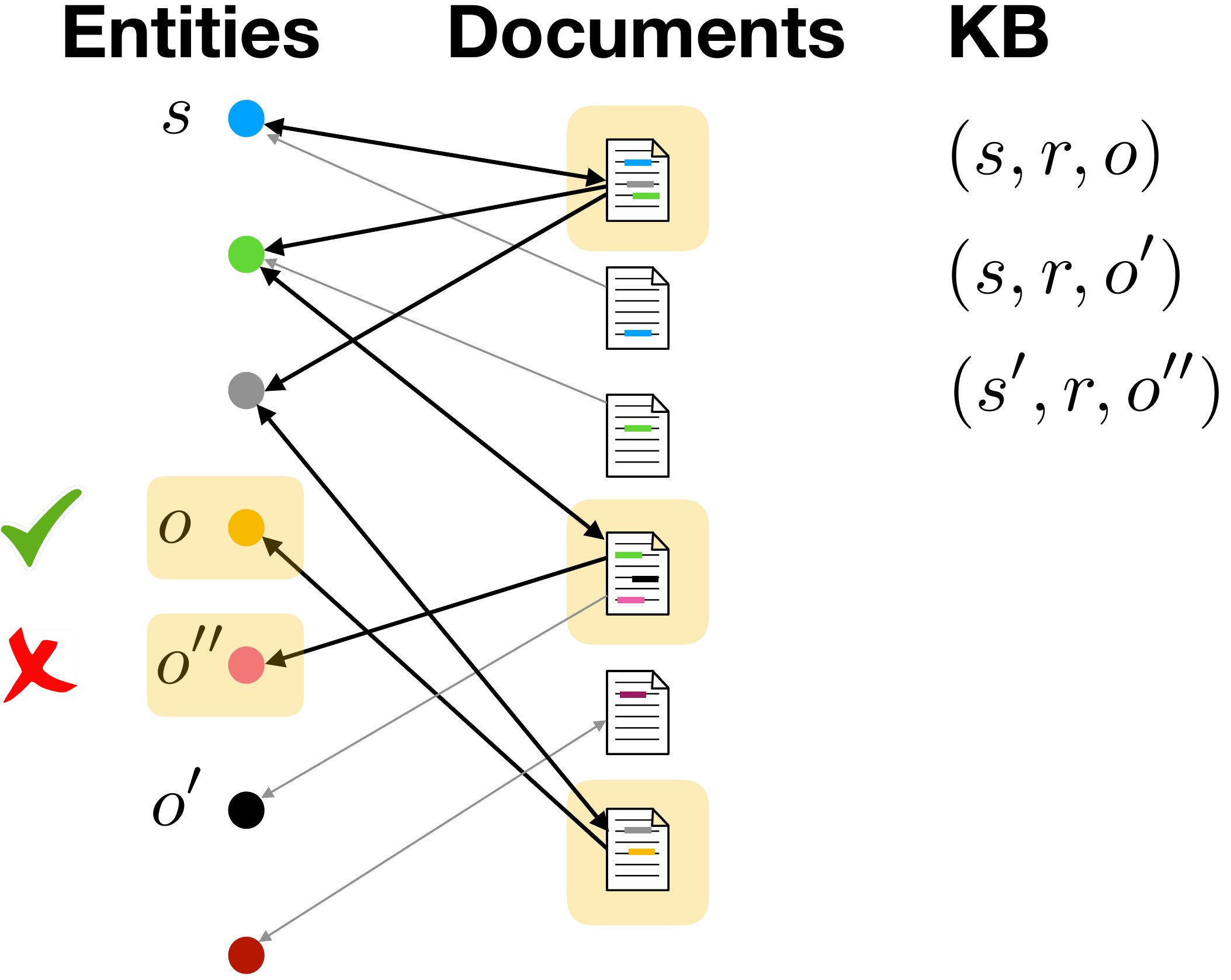}
    \caption{
        A bipartite graph given in paper\cite{ref_wikihop} connecting entities and documents mentioning them.
        Bold edges are those traversed for the first fact in the small KB on the right; 
        yellow highlighting indicates documents in $S_q$ and candidates in $C_q$.
        Check and cross indicate correct and false candidates.
    }
    \label{wikihop_pic_2}
\end{figure}

 To construct WIKIHOP, the authors collect \textit{(s, r, o)} triples - with subject entity $s$, relation $r$, and object entity $o$, from WIKIDATA\cite{ref_wikidata}.
%, which is widely used as source materials for building datasets. 
Then Wikipedia articles associated with the entities are added as candidate evidence documents $D$. The triple becomes a query after removing answer from it, that is, $q$ = \textit{(s, r, ?)} and  $a$=\textit{o}. To reach the goal of multi-hop reasoning, bipartite graphs are constructed for the help of corpus construction. As shown in Fig.\ref{wikihop_pic_2}, vertices on two sides respectively correspond to the entities and the documents from the Knowledge Base, and edges denote the entities appear in the corresponding documents. For a given \textit{(q,a)} pair, the answer candidates $C_q$ and support documents $Sq \in D$ are identified by traversing the bipartite graph using breadth-first search; the documents visited will become the support documents $Sq$.

%Then Wikipedia articles associated with the entities are added as candidate evidence documents $D$. The triple becomes a query after removing answer from it, in other word, $q$ = \textit{(s, r, ?)}  $a$=\textit{o}. Some further modifications as follows are required to reach the goal of multi-hop reasoning. 
%\par In the bipartite graph in Fig.\ref{wikihop_pic_2}, where vertices on two sides respectively correspond to the entities and the documents from the Knowledge Base, edges denote the entities appear in the corresponding documents. For a given \textit{(q,a)} pair, the candidates of answer $C_q$ and support documents $Sq \in D$ are identified by traversing the bipartite graph using breadth-first search; the documents visited will become the support documents $Sq$.
%\par The traversal starts with nodes belonging to the subject entity $s$, and ends with nodes that are type-consistent\footnote{A type-consistent entity of query $q$ refers to one that is observed as object in a fact with $r$ as relation type.} answers to $q$. When traversing the graph starting at $s$, several end points will be visited, and they are added to the candidate set $C_q$. 
\par  
Another dataset MEDHOP is constructed in the same way as WIKIHOP, with the focus on the medicine area.
Some basic statistics of WIKIHOP  and MEDHOP are shown in Table \ref{tbl:dataset_sizes} and Table \ref{tbl:dataset_minmaxmean}.  Table \ref{tbl:wikihop_qualitative} lists the proportions  of different types of answer samples, which indicates that to perform well on WIKIHOP, one system needs to be good at multi-step reasoning.

%wikihop 图片
% \begin{figure}
% \centering
% \includegraphics[width=0.6\textwidth]{wikihop_pic_1.png}
% \caption{}
% \label{wikihop_pic_1}
% \end{figure}
%第一张图是样例，换成新格式的表格了，如下
\begin{figure}
\setlength{\topsep}{0pt}
  \begin{framed}
    The Hanging Gardens, in {\color{blue} [Mumbai]}, also known as Pherozeshah Mehta Gardens, are terraced gardens ... They provide sunset views over the {\color{red} [Arabian Sea]} ...
 \end{framed}
 \begin{framed}
    {\color{blue} [Mumbai]} (also known as Bombay, the official name until 1995) is the capital city of the Indian state of Maharashtra. It is the most populous city in {\color{green} India} ... 
 \end{framed}
 \begin{framed}
    The {\color{red} [Arabian Sea]} is a region of the northern Indian Ocean bounded on the north by  Pakistan and Iran, on the west by northeastern Somalia and the Arabian Peninsula, and on the east by {\color{green} India} ... 
  \end{framed}
  \centering 
 \textbf{Question}: (Hanging gardens of Mumbai, country, ?) \\
 \textbf{Options}: \{Iran,{\color{green} India}, Pakistan, Somalia, ...\} 
  \caption{A sample of WIKIHOP quoted from \cite{ref_wikihop} which displays the necessity of multi-hop reasoning across several documents.}
  \label{wikihop_sample}
\end{figure}

% 描述过程的二部图，感觉不好删就留下来了
% \begin{figure}
% \centering
% \includegraphics[width=0.6\textwidth]{wikihop_pic_2.png}
% \caption{bipartite graph given in paper\cite{ref_wikihop}}
% \label{wikihop_pic_2}
% \end{figure}
% %

% WIKIHOP 的基本statistic，换成表格了
\begin{table}
    \centering
    \setlength{\tabcolsep}{7mm}{
    \begin{tabular}{lrrrr}
    \toprule
                    & Train     & Dev   & Test  & Total     \\
        \midrule
        WIKIHOP    & 43,738    & 5,129 & 2,451 & 51,318    \\
        MEDHOP     &  1,620    &   342 &   546 &  2,508     \\
    \bottomrule
    \end{tabular}}
    \caption{
        Dataset sizes of WIKIHOP and MedHop \cite{ref_wikihop}.
    }
    \label{tbl:dataset_sizes}
\end{table}
\begin{table}
    \centering
    \setlength{\tabcolsep}{6mm}{
    \begin{tabular}{lrrrr}
    \toprule
                            & min       & max   & avg   &  median       \\
        \midrule
        \# cand.  -- WH     & 2         & 79    & 19.8  & 14    \\
        \# docs.  -- WH     & 3         & 63    & 13.7  & 11    \\
        \# tok/doc -- WH & 4         & 2,046  & 100.4 & 91    \\
        \midrule
        \# cand.  -- MH     & 2         & 9     & 8.9   & 9      \\
        \# docs.  -- MH     & 5         & 64    & 36.4  & 29    \\
        \# tok/doc  -- MH   & 5         & 458    & 253.9  & 264    \\
        \bottomrule
    \end{tabular}}
    \caption{
        Corpus statistics of WIKIHOP and MedHop \cite{ref_wikihop}. WH: WikiHop; MH: MedHop.
    }
    \label{tbl:dataset_minmaxmean}
\end{table}

% sample的分析，用表格代替了图片
\begin{table}[t]
    \centering
    \setlength{\tabcolsep}{10mm}{
    \begin{tabular}{lr}
    \toprule
        Unique multi-step answer.  & 36\%                          \\
        Likely multi-step unique answer.                         &  9\%                          \\
        Multiple plausible answers.          & 15\%                          \\
        Ambiguity due to hypernymy.                              & 11\%                          \\
        Only single document required.           &  9\%                          \\
        \midrule
        Answer does not follow.                              & 12\%                          \\
        Wikidata/Wikipedia discrepancy.     &  8\%                          \\
        \bottomrule
    \end{tabular}}
    \caption{
       Qualitiative analysis of sampled answers of WIKIHOP\cite{ref_wikihop}
    }
    \label{tbl:wikihop_qualitative}
\end{table}

%%%%%%%%%%%%%%%%%%%%%%%%%%%
% 描述式的数据集的罗列
%%%%%%%%%%%%%%%%%%%%%%%%%%%
\subsection{Descriptive Answer Datasets}
\par Instead of text spans or entities obtained from candidate documents, descriptive answers  are whole, stand-alone sentences, which exhibit more fluency and integrity. In addition, in real world, many questions may not be answered simply by a text span or an entity. 
%, whereas most datasets described above all get caught in that. 
What's more, presenting answers with their supporting evidence and examples is preferred by human. So in light of these reasons, some descriptive answer datasets are released in recent years. 
% 总起来说一下，下面将介绍哪些数据集
Next we mainly introduce two of them in detail, namely MS MARCO and NarrativeQA.
% ***除了这两个，还有别的吗？？？

\paragraph{MS MARCO} MS MARCO (Microsoft MAchine Reading COmprehension) is a large  dataset released by Microsoft in 2016\cite{nguyen2016ms}. 
This dataset aims to address questions and documents in the real world.
%This dataset aims to address some major shortcomings of existing datasets, one of which is that problems conceived by crowdworkers from a given context differ a lot from those posed by real users\cite{nguyen2016ms}. What's more, questions and documents in real world can be messy or even self-conflicting, however the materials of many datasets above are of high quality, and this may lead to a catastrophe when a system trys to work on problematic real-world data. 
Sourced from real anonymized queries issued through Bing\footnote{www.bing.com} or Cortana\footnote{https://www.microsoft.com/en-us/cortana} and the corresponding searching results from Bing search engine, MS MARCO can well reproduce QA situations in real world. For each question in the dataset, a crowdworker is asked to answer it in the form of a complete sentence using passages provided by Bing. The unanswerable questions are also kept in the dataset for the purpose of encouraging one system to judge whether a question is answerable due to scanty or conflicting materials. The first version of MS MARCO released in 2016 has about 100k questions, and the latest version V2.1  released in 2018 has over 1,000k questions. Both are now available at \url{http://www.msmarco.org}.

% 下面这一段可以针对这些数据展开说一下
%% 已经添加了一些数据的讨论与描述
The dataset compositions of MS MARCO are shown in Table \ref{msmarco-composition}. And the distribution of different types of questions are shown in Table \ref{ms-query-percentage}. From this table, we can see that not all of them contain interrogatives, because the queries come from real users. We can also see that the interrogative "What" is contained in 34.96\% of the queries and description questions account for the major question type. Generally,  interrogative distribution  in questions shows reasonable diversity.

% 数据集的组成
\begin{table}
\centering
\begin{tabular}{r p{0.65\textwidth}} \\ \hline
\textbf{Field} & \textbf{Description} \\ \hline
Query & A question query issued to Bing.\\
Passages & Top 10 passages from Web documents as retrieved by Bing. The passages are presented in ranked order to human editors. The passage that the editor uses to compose the answer is annotated as is\_selected: 1.\\
Document URLs & URLs of the top ranked documents for the question from Bing. The passages are extracted from these documents. \\
Answer(s) & Answers composed by human editors for the question, automatically extracted passages and their corresponding documents. \\
Well Formed Answer(s) & Well-formed answer rewritten by human editors, and the original answer. \\
Segment & QA classification. E.g., {tallest mountain in south america} belongs to the ENTITY segment because the answer is an entity (Aconcagua). \\
\hline
\end{tabular}
\caption{The MS MARCO dataset composition \cite{nguyen2016ms}.}
\label{msmarco-composition}
\end{table}

% 数据集的答案分类，截图已经去掉了
% 文中有相应文字叙述，这里放上表格，更简洁些
\begin{table}
\label{ms-query-percentage}
\begin{center}
\setlength{\tabcolsep}{1pt}
\begin{tabular}{l l}
%\hline
{Question segment}&{ Percentage of question} \\
\hline
\textbf{Question types} & \\
YesNo & 7.46\% \\
What & 34.96\% \\
How & 16.8\% \\
Where & 3.46\% \\
When & 2.71\% \\
Why & 1.67\%  \\
Who & 3.33\% \\
Which & 1.79\% \\
Other & 27.83\% \\
\hline
\textbf{Question classification} & \\
Description & 53.12\% \\
Numeric & 26.12\% \\
Entity & 8.81\% \\
Location & 6.17\% \\
Person & 5.78\%  \\
\hline
\end{tabular}
\caption{Distribution of different question types in MS MARCO\cite{nguyen2016ms}}
\end{center}
\end{table}

% narrative 的一些newcommand
\newcommand{\examplea}[7]{%
\begin{figure}[tb]
\begin{tabular}{p{0.92\columnwidth}}
\toprule
\noindent%
\textbf{Title:} #1\\
\textbf{Question:} #2\\
\textbf{Answer:} #3\\
\textbf{Summary snippet:} #4\\
\textbf{Story snippet:} #5 \\
\bottomrule
\end{tabular}
\caption{#6}
\label{#7}
\end{figure}%
}
\newcommand{\ghostexample}{%
\examplea{Ghostbusters II}
{\small How is Oscar related to Dana?}
{\small her son}
{\small\dots Peter's former girlfriend Dana Barrett has had a son, Oscar\dots}
{\\
\small \centerline{\textit{DANA (setting the wheel brakes on the buggy)}}
\small \centerline{Thank you, Frank.  I'll get the hang of this eventually.}

\vspace{0.2cm}
\small She continues digging in her purse while Frank leans over the buggy and makes funny faces at the baby, OSCAR, a very cute nine-month old boy.

\vspace{0.2cm}
\small\centerline{\textit{FRANK (to the baby)}}
\small\centerline{Hiya, Oscar.  What do you say, slugger?}

\vspace{0.2cm}
\small\centerline{\textit{FRANK (to Dana)}}
\small That's a good-looking kid you got there, Ms.\ Barrett.}
{An example question-answer pair of NarrativeQA given in paper\cite{kovcisky2018narrativeqa}}
{narrative-example}}

\paragraph{NarrativeQA} NarrariveQA\cite{kovcisky2018narrativeqa} is another dataset with descriptive answers released by DeepMind and University of Oxford in 2017. 
NarrativeQA is specifically designed to examine how well a system can capture the underlying narrative elements to answer those questions which can not be answered by simple pattern recognition or global salience.
%Many existing datasets have the drawback that many questions can be answered by simple pattern recognition or global salience. 
%To overcome the drawback, NarrativeQA is specifically designed to examine how well a system can capture the underlying narrative elements rather than some superficial information. 
From an example of question-answer pair shown in Fig.\ref{narrative-example}, we can see  that  relatively high-level abstraction or reasoning is required to answer the question.
%\par The stories which are used to construct the question-answer pairs in NarrativeQA consist of books

The stories used in NarrativeQA consist of books from Project Gutenberg\footnote{http://www.gutenberg.org/} and movie scripts from relative websites\footnote{Mainly from http://www.imsdb.com/, and also from http://www.dailyscript.com/ and http://www.awesomefilm.com/.}. 
%Each story comes with a plot summary, which is finally provided to crowdworkers to create question-answer pairs.
Each story, as well as its plot summary, is finally provided to crowdworkers to create question-answer pairs. 
Because the crowdworkers never see the full text, it's less likely for them to create questions and answers solely based on localized context. 
The answers can be full sentences, which exhibit more artificial intelligence when asked about factual information\cite{kovcisky2018narrativeqa}.
%, or short phrases.
% when asked about ... short phrases这儿什么意思呢？？？

\ghostexample

Some basic statistics are shown in Table \ref{narrative_datasetstats}, and the distribution of different types of questions and answers are shown in Table \ref{narrative_first_token} and Table \ref{narrative_question_categories}. According to the original paper, less than 30\% of answers appear as text segments of the stories, which decreases the possibility of answering questions with simple skills for a system as before.
%for a system to solve questions with simple skills mentioned before.

% narrative 的图片

\begin{table}
\small
\footnotesize
\centering
\newcommand{\tokens}{tok.}
\setlength{\tabcolsep}{8mm}{
\begin{tabular}[t]{@{}llll@{}}
    \toprule
                                &  \textbf{train}   &  \textbf{valid}   &   \textbf{test}    \\
    \midrule
    \# documents                & 1,102    & 115      & 355       \\
    \dots\ books                & 548      & 58       & 177      \\
    \dots\ movie scripts        & 554      & 57       & 178       \\
    \# question--answer pairs    & 32,747   & 3,461    & 10,557    \\
    Avg. \#\tokens\ in summaries & 659      & 638      & 654       \\
    Max  \#\tokens\ in summaries & 1,161     & 1,189   & 1,148    \\
    Avg. \#\tokens\ in stories   & 62,528   & 62,743   & 57,780   \\
    Max  \#\tokens\ in stories   & 430,061  & 418,265  & 404,641    \\
    Avg. \#\tokens\ in questions & 9.83     & 9.69     & 9.85       \\
    Avg. \#\tokens\ in answers   & 4.73     & 4.60     & 4.72       \\
    \bottomrule
\end{tabular}}
\caption{NarrativeQA dataset statistics \cite{kovcisky2018narrativeqa}}
\label{narrative_datasetstats}

\vspace{1cm}

\begin{minipage}[t]{.45\linewidth}
\small
\footnotesize
\centering
\setlength{\tabcolsep}{8mm}{
\begin{tabular}[t]{@{}lr@{}}
    \toprule
    \textbf{First token} & \textbf{Frequency} \\
    \midrule
    What&	38.04\% \\
    Who&	23.37\% \\
    Why&	~9.78\% \\
    How&	~8.85\% \\
    Where&	~7.53\% \\
    Which&	~2.21\% \\
    How many/much& 1.80\% \\
    When&	~1.67\% \\
    In&	    ~1.19\% \\
    OTHER& ~5.57\% \\
    \bottomrule
\end{tabular}}
\caption{Frequency of first token of the question in the training set of NarrativeQA \cite{kovcisky2018narrativeqa}.}
\label{narrative_first_token}
\end{minipage}%
\hfill
\begin{minipage}[t]{.45\linewidth}
\small
\footnotesize
\centering
\setlength{\tabcolsep}{8mm}{
\begin{tabular}[t]{@{}lr@{}}
    \toprule
    \textbf{Category} & \textbf{Frequency} \\
    \midrule
Person	&30.54\%  \\
Description~~~	&24.50\%  \\
Location	&~9.73\%  \\
Why/reason	&~9.40\%  \\
How/method	&~8.05\%  \\
Event	&~4.36\%  \\
Entity	&~4.03\%  \\
Object	&~3.36\%  \\
Numeric	&~3.02\%  \\
Duration	&~1.68\%  \\
Relation	&~1.34\%  \\
\bottomrule
\end{tabular}}
\caption{Question categories on a sample of 300 questions from the validation set of NarrativeQA \cite{kovcisky2018narrativeqa}.}
\label{narrative_question_categories}
\end{minipage}
\end{table}

%%%%%%%%%%%%%%%%
%%%  总结各个数据集的表格，太长了。。
% 可以考虑放在附录中
%%%%%%%%%%%%%%%%%
% basic info of extractive and narrative
\begin{landscape}
\begin{ThreePartTable}
  \begin{TableNotes}
\item[a] <RougeL-Bleu1>,on Q\&A + Natural Langauge Generation Task.
\item[b] <Bleu1-Bleu4-Meteor-RougeL>.
  \end{TableNotes}

\setlength{\tabcolsep}{1mm}{
\begin{longtable}{c|cccccccc|cc}

% name of dataset
& \begin{tabular}{@{}c@{}}SQuAD\\(v1.1)  \end{tabular} 
& \begin{tabular}{@{}c@{}}SQuAD\\(v2.0)  \end{tabular} 
& \begin{tabular}{@{}c@{}}CNN\&\\Daily Mail \end{tabular} 
& CBT & NewsQA  & TriviaQA  & WIKIHOP & & MS MARCO & NarrativeQA
\\ \hline

% date
\begin{tabular}{@{}c@{}}Release \\ date\end{tabular}  & 2016 & 2018 & 2015 & 2015 & 2017 & 2017 & 2018 & & 2018(v2) & 2017
\\ \hline

% type
Type & extractive & extractive & extractive & extractive & extractive & extractive & extractive& & narrative & narrative \\ \hline

% domain
domain & Wikipedia  & Wikipedia  & News & books  & News & Trivia & Wikipedia& & Search Engine & Scripts
\\ \hline

%source 
\begin{tabular}{@{}c@{}}Question \\ source\end{tabular} & \begin{tabular}{@{}c@{}}crowd\\-sourced\end{tabular} & \begin{tabular}{@{}c@{}}crowd\\-sourced\end{tabular} & automatic & automatic & \begin{tabular}{@{}c@{}}crowd\\-sourced\end{tabular} & natural & automatic & 
& \begin{tabular}{@{}c@{}}query:natural\\ answer:automatic\end{tabular} 
& \begin{tabular}{@{}c@{}}crowd\\-sourced\end{tabular}
\\ \hline

% human perf
\begin{tabular}{@{}c@{}}Human\\ performance \end{tabular}
& \begin{tabular}{@{}c@{}}EM 82.3\\ F1 91.2\end{tabular} 
& \begin{tabular}{@{}c@{}}EM 86.8 \\ F1 89.5\end{tabular} 
& - 
& \begin{tabular}{@{}l@{}}NE 0.816\\ CN 81.6\\ VB 82.8\\ PR. 70.8\end{tabular} 
& \begin{tabular}{@{}l@{}}EM 46.5 \\ F1 74.9 \end{tabular} 
& \begin{tabular}{@{}l@{}}wiki-dom\\ 79.7 \\ web-dom\\ 75.4 \end{tabular}
& 85.0 
& & 63.21-53.03 \tnote{a}
& \begin{tabular}{@{}l@{}}44.43-19.65\\-24.14-57.02 \end{tabular} \tnote{b}
\\ \hline

SOTA 
& \begin{tabular}{@{}c@{}}EM 87.4\\ F1 93.2 \end{tabular} 
& \begin{tabular}{@{}c@{}}EM 85.1\\ F1 87.6 \end{tabular}
& \begin{tabular}{@{}c@{}}EM 76.9 \\ F1 79.6 \end{tabular}
& \begin{tabular}{@{}c@{}}NE 89.1 \\ CN 93.3 \end{tabular}
& \begin{tabular}{@{}c@{}}EM 42.8 \\ F1 56.1 \end{tabular} 
& \begin{tabular}{@{}c@{}}wiki-dom\\ 67.3 \\ web-dom\\ 68.7\end{tabular} 
& 71.2 
&
& 49.61-50.13
& \begin{tabular}{@{}c@{}}44.35-27.61\\ -21.80-44.69\end{tabular} 
\\ \hline

\begin{tabular}{@{}c@{}}Contain\\unanswerable\\question\end{tabular} 
& \xmark & \cmark & \xmark & \xmark & \cmark & \xmark & \xmark & & \cmark & \xmark 
\\ \hline

\caption{basic info of all Extractive and Narrative datasets}\\
\insertTableNotes
\end{longtable}}
\end{ThreePartTable}
\end{landscape}

% stat info of extractive and narrative
\begin{landscape}
\begin{ThreePartTable}
  \begin{TableNotes} 
\item[*] Default use word number when calculating length unless specified. 
\item[$\dagger$] The statistics with $\dagger$ are counted by ourselves. Unless specified other statistics come from corresponding original papers.
\item[?] Corresponding data is unavailable.
\item[a] Wikipedia articles, <train-dev-test>.
\item[b] Months of news, <train-dev-test>.
\item[c] Books number, <train-dev-test>.
\item[d] News articles.
\item[e] Full Web Documents
\item[f] Stories, <train-dev-test>.
\item[g] Paragraphs,<train-dev-test>.
\item[h] Passages.
\item[i] Result from \cite{chen2016thorough}.
\item[j] Anonymised version, the answer is an entity marker.
\end{TableNotes}
  
%
%\centering
\begin{longtable}{c|cccccccc|cc}
%\hline

% name of dataset
& \begin{tabular}{@{}c@{}}SQuAD \\(v1.1)  \end{tabular} 
& \begin{tabular}{@{}c@{}}SQuAD\\(v2.0)  \end{tabular} 
& \begin{tabular}{@{}c@{}}CNN\&\\Daily Mail \end{tabular} 
& CBT & NewsQA  & TriviaQA  & WIKIHOP & & MS MARCO & NarrativeQA
\\ \hline

% raw document
\begin{tabular}{@{}c@{}}Raw\\ document\end{tabular}
& \begin{tabular}{@{}c@{}}442-\\48-\\46 \tnote{a}\end{tabular}  
& \begin{tabular}{@{}c@{}}442-\\ 35-\\ 28 \tnote{a}\end{tabular}
& \begin{tabular}{@{}c@{}}95-1-1\\\&\\56-1-1\tnote{b}\end{tabular} 
& \begin{tabular}{@{}c@{}} 98-\\ 5-\\ 5 \tnote{c}\end{tabular}
& 	12,744 \tnote{d}
& -
& -
& 
& 3,563,535 \tnote{e} 
& \begin{tabular}{@{}c@{}}1,102-\\115-\\355 \tnote{f}\end{tabular}
\\ \hline

% document 
\begin{tabular}{@{}c@{}}Document\\number\end{tabular} 
& \begin{tabular}{@{}c@{}}18896-\\2067-\\? \tnote{g}\end{tabular}$\dagger$ 
& \begin{tabular}{@{}c@{}}19035-\\1204-\\? \tnote{g}\end{tabular}$\dagger$
& \begin{tabular}{@{}c@{}}90226-\\1220-\\1093 \\\&\\ 196961-\\12148-\\10397\tnote{d}\end{tabular} 
& \begin{tabular}{@{}c@{}}669343-\\8000-\\10000 \tnote{g}\end{tabular}
& 12,744	\tnote{d}
& 662,659 \tnote{g}
& \begin{tabular}{@{}c@{}}598103-\\74741-\\? \tnote{g}\end{tabular}$\dagger$
& 
& \begin{tabular}{@{}c@{}}8069749-\\1008985-\\1008943 \tnote{h}\end{tabular}$\dagger$
& \begin{tabular}{@{}c@{}}1102-\\115-\\355 \tnote{f}\end{tabular}
\\ \hline

% ave len of document
\begin{tabular}{@{}c@{}}Average \\length of \\ document \tnote{1} \end{tabular}
& \begin{tabular}{@{}c@{}}116.6-\\122.8-\\?\end{tabular}$\dagger$
& \begin{tabular}{@{}c@{}}116.6-\\126.6-\\?\end{tabular}$\dagger$
& \begin{tabular}{@{}c@{}}762-\\763-\\716\\\& \\813-\\774-\\780\end{tabular} 
& \begin{tabular}{@{}c@{}}465-\\435-\\445\end{tabular}
& 616
& 2,895
& \begin{tabular}{@{}c@{}}85.42-\\85.01-\\?\end{tabular}$\dagger$
& 
& \begin{tabular}{@{}c@{}}56.49-\\53.04-\\53.05\end{tabular}$\dagger$
& \begin{tabular}{@{}c@{}}62528-\\62743-\\57780\end{tabular}
\\ \hline

% query number
\begin{tabular}{@{}c@{}}Query\\number\end{tabular}
& \begin{tabular}{@{}c@{}}87599-\\10570-\\9533\end{tabular}	
& \begin{tabular}{@{}c@{}}130319-\\11873-\\8862\end{tabular}
& \begin{tabular}{@{}c@{}}380298-\\3924-\\3198\\\&\\879450-\\64835-\\53182\end{tabular}  
& \begin{tabular}{@{}c@{}}669343-\\8000-\\10000\end{tabular}
& 119,633
& 95,956
& \begin{tabular}{@{}c@{}}43,738-\\5,129-\\2,451\end{tabular}
& 
& \begin{tabular}{@{}c@{}} 808731-\\101093-\\101092\end{tabular}$\dagger$
& \begin{tabular}{@{}c@{}} 32747-\\3461-\\10557\end{tabular}
\\ \hline 

% ave len query
\begin{tabular}{@{}c@{}}Average \\length of \\ query\end{tabular}
& \begin{tabular}{@{}c@{}} 10.1-\\10.2-\\?\end{tabular}$\dagger$
& \begin{tabular}{@{}c@{}} 9.89-\\10.02-\\?\end{tabular}$\dagger$
& \begin{tabular}{@{}c@{}} 12.5\\\&\\ 14.3 \tnote{i}\end{tabular}
& \begin{tabular}{@{}c@{}} 31-\\27-\\29\end{tabular}	
& 6.77	$\dagger$
& 14	
& \begin{tabular}{@{}c@{}} 3.42-\\3.42-\\?\end{tabular}$\dagger$
& 
& \begin{tabular}{@{}c@{}} 6.37-\\6.41-\\6.40\end{tabular}$\dagger$
& \begin{tabular}{@{}c@{}} 9.83-\\9.69-\\9.85\end{tabular}
\\ \hline

% ave answer len
\begin{tabular}{@{}c@{}}Average \\length of \\ answer\end{tabular}
& \begin{tabular}{@{}c@{}}3.16-\\2.91-\\?\end{tabular}	$\dagger$
& \begin{tabular}{@{}c@{}}3.16-\\3.06-\\?\end{tabular}$\dagger$
& 1	\tnote{j}
& 1	
& 4.13	
& 1.68	$\dagger$
& \begin{tabular}{@{}c@{}}1.79-\\1.73-\\?\end{tabular}	$\dagger$
& 
& \begin{tabular}{@{}c@{}} 9.21-\\ 9.65-\\?\end{tabular}$\dagger$
& \begin{tabular}{@{}c@{}} 4.73-\\ 4.60-\\ 4.72\end{tabular}
\\ \hline

\caption{Statistics infomation of all Extractive and Narrative datasets.}\\
\insertTableNotes
\end{longtable}
\end{ThreePartTable}
\end{landscape}

\begin{table}
\begin{minipage}{\textwidth}
\setlength{\tabcolsep}{1mm}{
\begin{tabular}{  p{0.16\textwidth} |  p{0.11\textwidth}   p{0.11\textwidth}   p{0.11\textwidth}   p{0.12\textwidth}  p{0.11\textwidth}   p{0.11\textwidth}  }

	 & RACE  & CLOTH  & MCTest  & MCScript  & ARC  & CoQA  
	 \\ \hline

	\begin{tabular}{@{}l@{}}Release \\date \end{tabular}  & 2017 & 2017 & 2013 & 2018 & 2018 & 2018 
	\\ \hline

	Type & multiple choice & multiple choice & multiple choice & multiple choice & multiple choice & multiple choice 
	\\ \hline

	Domain & exam & exam 
	& \begin{tabular}{@{}l@{}}Fiction\\ stories\end{tabular} 
	& \begin{tabular}{@{}l@{}}Script \\scenarios\end{tabular} 
	& science & Wide\footnote{Children's Stories, Literature, Mid/High School Exams, News, Wikipedia, Science, Reddit} 
	
 \\ \hline
 
	\begin{tabular}{@{}c@{}}Question \\ source\end{tabular}  & natural & natural & \begin{tabular}{@{}c@{}}crowd\\-sourced\end{tabular} & \begin{tabular}{@{}c@{}}crowd\\-sourced\end{tabular} & natural & \begin{tabular}{@{}c@{}}crowd\\-sourced\end{tabular}
	
	\\ \hline

	\begin{tabular}{@{}l@{}}Human\\performance \end{tabular} 
& \begin{tabular}{@{}l@{}}95.4-\\94.2 \end{tabular} \footnote{RACE-M - RACE-H}
& \begin{tabular}{@{}l@{}}85.9-\\89.7-\\84.5  \end{tabular} \footnote{total-middle-high}
& \begin{tabular}{@{}l@{}}97.7-\\96.9 \end{tabular} \footnote{MC160-MC500}
& 98.2 
& - 
& \begin{tabular}{@{}l@{}}89.4-\\87.4 \end{tabular} \footnote{in domain-out of domain}

 \\ \hline
	SOTA & 
\begin{tabular}{@{}l@{}}73.4-\\68.1 \end{tabular}\footnote{RACE-M-RACE-H}
 & \begin{tabular}{@{}l@{}}0.860-\\0.887-\\0.850 \end{tabular}
 & \begin{tabular}{@{}l@{}} 81.7-\\ 82.0 \end{tabular}
 & 84.84
 & 44.62
 & \begin{tabular}{@{}l@{}} 87.5-\\ 85.3 \end{tabular}
 
 \\ \hline

	\begin{tabular}{@{}c@{}}Contain\\unanswerable\\question\end{tabular}  & \xmark & \xmark & \xmark & \xmark & \xmark & \cmark \\ \hline

	\begin{tabular}{@{}c@{}}Test common \\sense \\specifically\end{tabular}& \xmark & \xmark & \xmark & \cmark & \cmark & \xmark 
	\\ \hline
	
	 &  &  &  &  &  &  \\ \hline

	\begin{tabular}{@{}c@{}}Raw\\ document\end{tabular} & - & - & - & 110\footnote{scenarios}  & 14M\footnote{science-related sentences}
 & - 
 \\ \hline

	\begin{tabular}{@{}c@{}}Document\\number\end{tabular}  
	& \begin{tabular}{@{}c@{}} 25,137-\\ 1,389-\\ 1,407\end{tabular}
	& \begin{tabular}{@{}c@{}}5513-\\805-\\813\end{tabular}
 & \begin{tabular}{@{}c@{}} 160-\\ 500\footnote{stories} \end{tabular}
 & \begin{tabular}{@{}c@{}} 1470-\\219-\\ 430\footnote{texts}\end{tabular} 
 & - 
 & 8,399 \footnote{Passages}
 \\ \hline

	\begin{tabular}{@{}c@{}}Average \\length of \\ document \tnote{1} \end{tabular} 
	& 321.9 
	& 313.16 
	& \begin{tabular}{@{}c@{}} 204-\\ 212\end{tabular}
 & 196 & - & 271 
 \\ \hline
 
	\begin{tabular}{@{}c@{}}Query\\number\end{tabular} 
	& \begin{tabular}{@{}c@{}} 87,866-\\ 4,887-\\4,934\\\end{tabular}
	& \begin{tabular}{@{}c@{}} 76850-\\ 11067-\\11516\\\end{tabular}
	& \begin{tabular}{@{}c@{}} 640-\\2000\end{tabular}
	& \begin{tabular}{@{}c@{}} 9731-\\ 1411-\\2797\end{tabular}
	& \begin{tabular}{@{}c@{}} 3370-\\ 869-\\ 3548\end{tabular}
& 127k 
\\ \hline

	\begin{tabular}{@{}c@{}}Average \\length of \\ query\end{tabular} & 10 & - & \begin{tabular}{@{}c@{}} 8.0-\\ 7.7\end{tabular} & 7.8 & 20.4 & 5.5 \\ \hline
	\begin{tabular}{@{}c@{}}Average \\length of \\ answer\end{tabular} & 5.3 & 1 & \begin{tabular}{@{}c@{}} 3.4-\\ 3.4\end{tabular} & 3.6 & 4.1 & 2.7 \\ \hline
\end{tabular}}
\end{minipage}
\caption{Basic information and statistics of all Multiple-choice datasets.}
\end{table}

\subsection{Multiple-choice}
%While descriptive answer datasets can achieve more fluency, it is more difficult to evaluate  system performance precisely and objectively. 
Datasets with descriptive answers are relatively difficult to evaluate the system performance precisely and objectively. 
Nevertheless, multiple-choice question, which has long been used for testing students reading comprehension ability, can be objectively gradable. Generally, this kind of questions can extensively examine one's reasoning skills, including simple pattern recognition, clausal inference and multiple-sentence reasoning, of a given passage. 
In light of this, many datasets in this format are released and listed as follows.

\paragraph{MCTest} 
%MCTest\cite{richardson2013mctest}, a high-quality dataset consists of 500 stories and 2000 questions about fiction stories, is in the same format as RACE whereas released much earlier in 2013 by Microsoft. 
MCTest\cite{richardson2013mctest}, a high-quality dataset consisting of 500 stories and 2000 questions about fiction stories, was  released  in 2013 by Microsoft with the same format as RACE.
Targeting at 7-year-old children, passages and questions used in MCTest are quite easy and understandable, which reduces the world knowledge requisite. 
%The stories are fictional, and the answers usually can only be found in the story. 
For MCTest, many answers can only be found in the story, since the stories are fictional.  
The main drawback of MCTest is that its size  is too small to train a well-performed model. A sample of MCTest is shown in Fig.\ref{mctest_sample}.

\begin{figure}
  \begin{framed}
     James the Turtle was always getting in trouble. Sometimes he'd reach into the freezer and empty out all the food. Other times he'd sled on the deck and get a splinter. His aunt Jane tried as hard as she could to keep him out of trouble, but he was sneaky and got into lots of trouble behind her back. One day, James thought he would go into town and see what kind of trouble he could get into. He went to the grocery store and pulled all the pudding off the shelves and ate two jars. Then he walked to the fast food restaurant and ordered 15 bags of fries. He did- n't pay, and instead headed home. His aunt was waiting for him in his room. She told James that she loved him, but he would have to start acting like a well-behaved turtle.After about a month, and after getting into lots of trouble, James finally made up his mind to be a better turtle. 
    \\
    \\
\par (1) What is the name of the trouble making turtle? \\
(A) Fries (B) Pudding (C) James (D) Jane 
\\
\par (2) What did James pull off of the shelves in the grocery store?\\ 
(A) pudding (B) fries (C) food (D) splinters  
\\
\par (3) Where did James go after he went to the grocery store?\\ (A) his deck (B) his freezer  (C) a fast food restaurant (D) his room 
\\
\par (4) What did James do after he ordered the fries? \\
(A) went to the grocery store (B) went home without paying\\ (C) ate them  (D) made up his mind to be a better turtle 
 \end{framed}
  \caption{A sample of MCTest given in paper\cite{richardson2013mctest}}
  \label{mctest_sample}
\end{figure}

%%%%%%%

\paragraph{RACE} RACE\cite{lai2017large} contains 27,933 passages and 97,687 questions that are collected from English exams for middle and high school Chinese students. Considering that those passages and questions are specifically designed by English teachers and experts to evaluate reading comprehension ability of students, this dataset is promising in developing and testing MRC systems.
\par Because the questions are created with high quality  by human experts, there are few noises in RACE. What's more, passages in RACE cover a wide range of topics, overcoming the topic bias problem that commonly exists in other datasets (like news articles for CNN/Daily Mail\cite{ref_cnn} and Wikipedia articles for SQuAD\cite{ref_squad}).
\par A sample of RACE is shown in Table \ref{race_pic_1}. The dataset firstly provides students/systems with a passage to read, then presents several  questions with 4 candidate answers. Words in the questions and candidate answers may not appear in the passage, so simple context-matching techniques will not aid as much as in other datasets. Analysis in the paper\cite{lai2017large} shows that reasoning skill is indispensable to answering most questions of RACE correctly.

\par 
%RACE is divided into two subsets, namely RACE-M and RACE-H, for the big difficulty gap between exams of middle school and that of high school.
RACE is divided into two subsets, namely RACE-M and RACE-H, for middle school and high school respectively.
Some basic statistics of RACE is given in Table \ref{race_pic_2} and Table \ref{race_pic_3}. Distributions of different reasoning types required to answer certain questions are illustrated in Table \ref{race_pic_4},  denoting that over  half of the questions in RACE requires \textit{Reasoning skill}.

%% race picture 

% 这个很长的table是RACE样例
\begin{table*}
	\centering
	\fbox{\begin{minipage}[t]{0.9\textwidth}
{\small
{\bf Passage:}

In a small village in England about 150 years ago, a mail coach was standing on the street. It didn't come to that village often. People had to pay a lot to get a letter. The person who sent the letter didn't have to pay the postage, while the receiver had to. 

``Here's a letter for Miss Alice Brown," said the mailman.

`` I'm  Alice Brown," a girl of about 18 said in a low voice.

Alice looked at the envelope for a minute, and then handed it back to the mailman.

``I'm sorry I can't take it, I don't have enough money to pay it", she said.

A gentleman standing around were very sorry for her. Then he came up and paid the postage for her.

When the gentleman gave the letter to her, she said with a smile, `` Thank you very much, This letter is from Tom. I'm going to marry him. He went to London to look for work. I've waited a long time for this letter, but now I don't need it, there is nothing in it."

``Really? How do you know that?" the gentleman said in surprise.

``He told me that he would put some signs on the envelope. Look, sir, this cross in the corner means that he is well and this circle means he has found work. That's good news."

The gentleman was Sir Rowland Hill. He didn't forgot Alice and her letter.

``The postage to be paid by the receiver has to be changed," he said to himself and had a good plan.

``The postage has to be much lower, what about a penny? And the person who sends the letter pays the postage. He has to buy a stamp and put it on the envelope." he said .
The government accepted his plan. Then the first stamp was put out in 1840. It was called the ``Penny Black". It had a picture of the Queen on it.

\vspace{1ex}

{\bf Questions:}
\begin{multicols}{2}
1): The first postage stamp was made \_.

A. in England
B. in America
C. by Alice
D. in 1910

\vspace{1ex}
2): The girl handed the letter back to the mailman because \_ .

A. she didn't know whose letter it was

B. she had no money to pay the postage

C. she received the letter but she didn't want to open it

D. she had already known what was written in the letter

\vspace{1ex}
3): We can know from Alice's words that \_ .

A. Tom had told her what the signs meant before leaving

B. Alice was clever and could guess the meaning of the signs

C. Alice had put the signs on the envelope herself

D. Tom had put the signs as Alice had told him to

\columnbreak

4): The idea of using stamps was thought of by \_ .

A. the government

B. Sir Rowland Hill

C. Alice Brown

D. Tom

\vspace{1ex}

5): From the passage we know the high postage made \_ .

A. people never send each other letters

B. lovers almost lose every touch with each other

C. people try their best to avoid paying it

D. receivers refuse to pay the coming letters

\vspace{1ex}

{\bf Answer:} ADABC
\end{multicols}
}
\end{minipage}}
\caption{A sample of RACE quoted from \cite{lai2017large}.}
\label{race_pic_1}
\end{table*}

% 第二第三个图是基本的stat，换成了表格
\begin{table*}
\centering
\footnotesize
\setlength{\tabcolsep}{1mm}{
\begin{tabular}{l|ccc|ccc|cccc}
\hline
Dataset & \multicolumn{3}{c|}{RACE-M} & \multicolumn{3}{c|}{RACE-H} & \multicolumn{4}{c}{RACE} \\ \hline
Subset                 & Train & Dev & Test & Train & Dev & Test & Train & Dev & Test & All    \\ \hline
\# passages        & 6,409    & 368   & 362    & 18,728   & 1,021       & 1,045  &25,137 &1,389& 1,407 & 27,933  \\
\# questions       & 25,421   & 1,436  & 1,436   & 62,445   & 3,451       & 3,498  &87,866 &4,887 &4,934 & 97,687  \\
\hline
\end{tabular}}
\caption{The basic statistics of the training, development and test sets of RACE-M,RACE-H and RACE\cite{lai2017large}}
\label{race_pic_2}
\end{table*}

\begin{table}[ht]
\centering
\setlength{\tabcolsep}{8mm}{
\begin{tabular}{l|c|c|c} \hline
Dataset & RACE-M & RACE-H & RACE \\ \hline
Passage Len  & 231.1               & 353.1               & 321.9  \\
Question Len & 9.0                 & 10.4               & 10.0    \\
Option Len & 3.9                 & 5.8                 & 5.3    \\
Vocab size        & 32,811               & 125,120               & 136,629 \\
\hline
\end{tabular}}
\caption{Statistics of RACE where Len denotes length and Vocab denotes Vocabulary \cite{lai2017large}.}
\label{race_pic_3}
\end{table}

%RACE 的第四个table是回答问题所需的能力
\begin{table*}
\centering
\setlength{\tabcolsep}{1mm}{
\begin{tabular}{l|cccccc}
\hline
Dataset       & RACE-M  & RACE-H  & RACE  & CNN & SQUAD   & NEWSQA  \\
\hline
Word Matching & 29.4\% & 11.3\% & 15.8\% & 13.0\%$^\dagger$ & 39.8\%* & 32.7\%* \\
Paraphrasing   & 14.8\% & 20.6\% & 19.2\% & 41.0\%$^\dagger$ & 34.3\%* & 27.0\%* \\
Single-Sentence Reasoning & 31.3\% & 34.1\% &33.4\% & 19.0\%$^\dagger$& 8.6\%* & 13.2\%*\\
Multi-Sentence Reasoning  & 22.6\% & 26.9\% & 25.8\% & 2.0\%$^\dagger$& 11.9\%* & 20.7\%* \\
Ambiguous/Insufficient     & 1.8\%  & 7.1\%  & 5.8\%  & 25.0\%$^\dagger$& 5.4\%*  & 6.4\%*  \\
\hline
\end{tabular}}
%\caption{Statistic information from paper\cite{lai2017large} about Reasoning type in different datasets. * denotes the numbers coming from \cite{ref_newsqa} based on 1000 samples per dataset, and numbers with $\dagger$ come from \cite{chen2016thorough}.
\caption{Distribution of reasoning type in RACE\cite{lai2017large} and other  datasets. * denotes quoting \cite{ref_newsqa} based on 1000 samples per dataset, and $\dagger$ quoting \cite{chen2016thorough}.}
\label{race_pic_4}
\end{table*}

%%%%%%%% CLOTH

\paragraph{CLOTH} 
CLOTH (CLOze test by TeacHers)~\cite{xie2017cloth} was constructed with the format of cloze questions. It is also composed of English tests for Chinese middle school and high school. One example is shown in Table \ref{sample-cloth}.
%CLOTH (CLOze test by TeacHers)~\cite{xie2017cloth} was constructed according to English cloze tests of Chinese middle school and high school, aiming at comprehensively evaluating students reading comprehension ability. One example is shown in Table \ref{sample-cloth}.
In CLOTH, the missing blanks in the questions were carefully designed by teachers to test different aspects of language knowledge.
The  candidate answers usually have subtle differences,  making the questions difficult to answer even for human.
%These two features distinguish CLOTH dataset from previous ones, which is either too small to train strong models or generated without specific emphasis on avoiding ambiguity and testing language ability.
%Carefully designed by teachers to evaluate students reading comprehension ability, the missing blanks of those questions are chosen to test certain aspects of language knowledge, and the candidates  usually have subtle differences, which make those questions difficult to answer even for human. These two features distinguish CLOTH dataset from previous ones, which is either too small to train strong models or generated without specific emphasis on avoiding ambiguity and testing language ability. 
Similar to RACE, CLOTH is also divided into two parts: CLOTH-M  for middle school and CLOTH-H for high school ones. Some basic statistics of this corpus are shown in Table \ref{stat_cloth}.
%Due to the difficulty difference of questions between middle school and high school, CLOTH is divided into two subsets, CLOTH-M for middle school questions and CLOTH-H for high school ones. Some basic statistics are shown in Table \ref{stat_cloth}.

Through experiments on CLOTH, the authors came to the conclusion that the performance gap between human and a system mainly results from the ability of using a long-term context~\cite{xie2017cloth}, or multiple-sentence reasoning. 
%the models’ inability to use a long-term context~\cite{xie2017cloth}, or, multiple-sentence reasoning. 
%Different reference skills\footnote{GM stands for grammar, STR stands for short-term reasoning, MP stands for matching or paraphrasing and LTR stands for long-term reasoning.} and their proportion are shown in Fig.\ref{cloth_pic_3}.
%

% cloth的样例
\begin{table}[th]
{\footnotesize
{\bf Passage:} 
Nancy had just got a job as a secretary in a company. Monday was the first day she went to work, so she was very \_1\_ and arrived early. She \_2\_ the door open and found nobody there. "I am the \_3\_ to arrive." She thought and came to her desk. She was surprised to find a bunch of \_4\_ on it. They were fresh. She \_5\_ them and they were sweet. She looked around for a \_6\_ to put them in. "Somebody has sent me flowers the very first day!" she thought \_7\_ . " But who could it be?" she began to \_8\_ . The day passed quickly and Nancy did everything with \_9\_ interest. For the following days of the \_10\_ , the first thing Nancy did was to change water for the followers and then set about her work. 

Then came another Monday. \_11\_ she came near her desk she was overjoyed to see a(n) \_12\_ bunch of flowers there. She quickly put them in the vase, \_13\_ the old ones. The same thing happened again the next Monday. Nancy began to think of ways to find out the \_14\_ . On Tuesday afternoon, she was sent to hand in a plan to the \_15\_ . She waited for his directives at his secretary's \_16\_ . She happened to see on the desk a half-opened notebook, which \_17\_ : "In order to keep the secretaries in high spirits, the company has decided that every Monday morning a bunch of fresh flowers should be put on each secretary’s desk." Later, she was told that their general manager was a business management psychologist.

\vspace{1ex}

{\bf Questions:}

\begin{tabular}{l@{\hskip 0.05in}l@{\hskip 0.05in}l@{\hskip 0.05in}l@{\hskip 0.05in}l}
1. & A. depressed&B. encouraged&\textbf{C. excited}&D. surprised \\
2. & A. turned&\textbf{B. pushed}&C. knocked&D. forced \\
3. & A. last&B. second&C. third&\textbf{D. first} \\
4. & A. keys&B. grapes&\textbf{C. flowers}&D. bananas \\
5. & \textbf{A. smelled}&B. ate&C. took&D. held \\
6. & \textbf{A. vase}&B. room&C. glass&D. bottle \\
7. & A. angrily&B. quietly&C. strangely&\textbf{D. happily }\\
8. & A. seek&\textbf{B. wonder}&C. work&D. ask \\
9. & A. low&B. little&\textbf{C. great}&D. general \\
10. & A. month&B. period&C. year&\textbf{D. week} \\
11. & A. Unless&\textbf{B. When}&C. Since&D. Before \\
12. & A. old&B. red&C. blue&\textbf{D. new} \\
13. & A. covering&B. demanding&\textbf{C. replacing}&D. forbidding \\
14. & \textbf{A. sender}&B. receiver&C. secretary&D. waiter \\
15. & A. assistant&B. colleague&C. employee&\textbf{D. manager} \\
16. & A. notebook&\textbf{B. desk}&C. office&D. house \\
17. &\textbf{A. said}&B. written&C. printed&D. signed \\
\end{tabular}

\begin{tabular}{lllll}
\end{tabular}
}

\caption{A Sample passage of CLOTH \cite{xie2017cloth}. Bold faces highlight the correct answers. There is only one best answer among four candidates, although several candidates may seem correct.}
\label{sample-cloth}
\end{table}

% cloth的stat
\begin{table*}[ht]
\centering
\footnotesize
\setlength{\tabcolsep}{1mm}{
\begin{tabular}{l|ccc|ccc|ccc}
\toprule
\multirow{2}{*}{Dataset} & \multicolumn{3}{c|}{CLOTH-M} & \multicolumn{3}{c|}{CLOTH-H} & \multicolumn{3}{c}{CLOTH} \\ 

                 & Train & Dev  & Test & Train & Dev  & Test & Train & Dev   & Test      \\ \midrule 
                 \# passages            & 2,341  & 355  & 335  & 3,172  & 450  & 478  & 5,513  & 805   & 813    \\ 

\# questions           & 22,056 & 3,273 & 3,198 & 54,794 & 7,794 & 8,318 & 76,850 & 11,067 & 11,516  \\ 

Vocab. size         & & {15,096} & & & {32,212} & & & {37,235} & \\ 
\midrule

Avg. \# sentence            & & {16.26} & & & 18.92 & & & 17.79 & \\ 
Avg. \# words               &  & 242.88 &  & & 365.1 & & & 313.16 & \\ 
\bottomrule
\end{tabular}}
\caption{The statistics of the training, development and test sets of CLOTH and two subsets from paper\cite{xie2017cloth}.}
\label{stat_cloth} 
\end{table*}

\begin{figure}
	\centering
	\includegraphics[width=0.6\columnwidth]{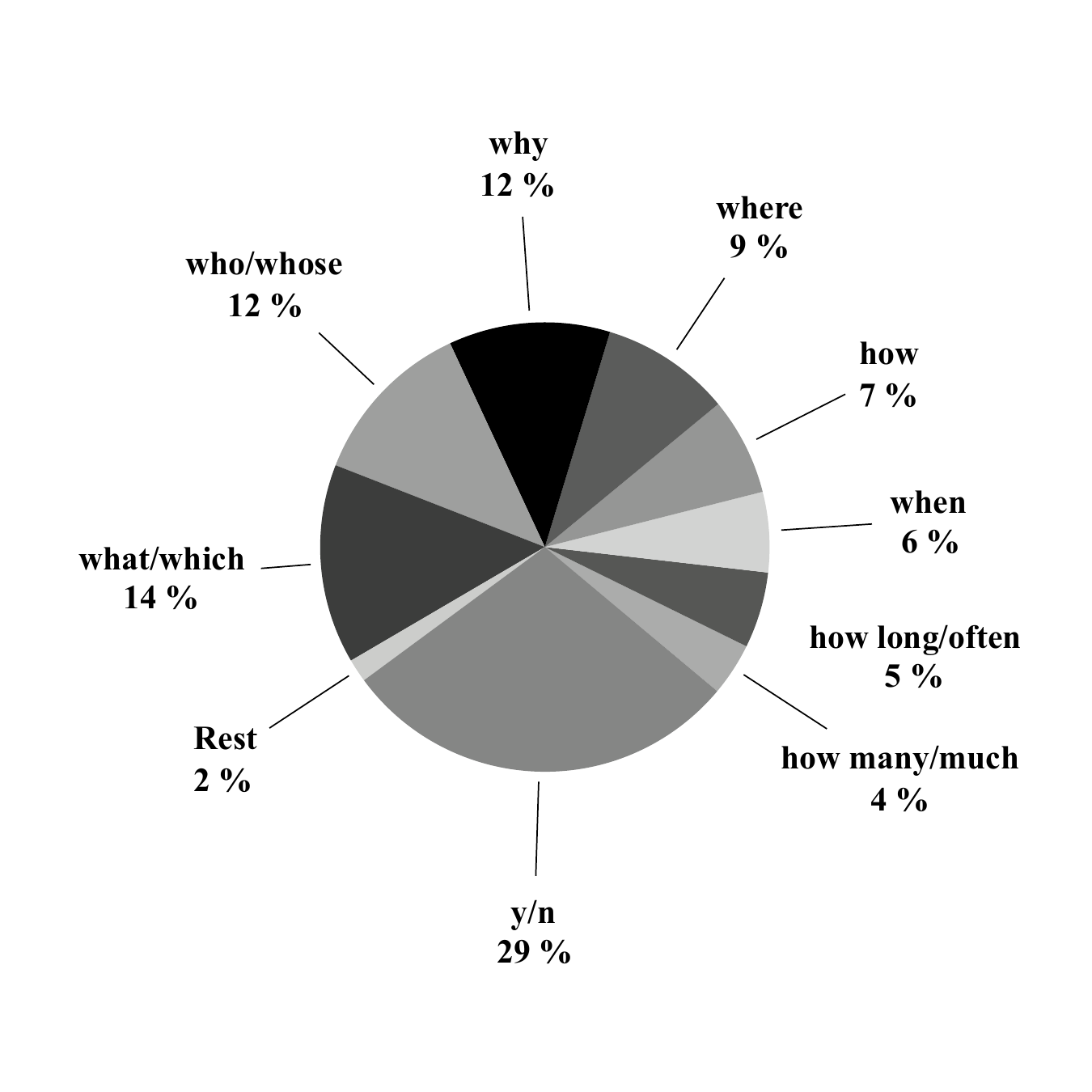}
	\caption{Distribution of question types in MCScript\cite{ostermann2018mcscript}.}
	\label{mcscript_pic_2}
\end{figure}

\paragraph{MCScript} MCScript\cite{ostermann2018mcscript} focuses on questions that need reasoning using commonsense knowledge. % to solve. 
Released in March 2018,
%this new dataset provided stories describing people’s daily activities, in which ambiguity and implicitness that can be resolved easily by commonsense exist, with crowdworkers to generate questions.
this new dataset provides stories describing people's daily activities, in which ambiguity and implicitness can be resolved easily by commonsense, with crowdworkers to generate questions.
The correct answers to the questions may not appear in the given text, as is shown in the examples in Fig.\ref{mcscript_pic_1}. It consists of about 2.1K texts and 14K questions. 
%And it was suggested by statistical analysis that 27.4\% of all the questions require commonsense knowledge to answer, so this dataset can literally examine systems’ commonsense inference ability.
According to statistical analysis, 27.4\% of all the questions in MCScript require commonsense knowledge to answer. Thus, this dataset can literally examine systems' commonsense inference ability.
All questions in the dataset are answerable. The distribution of the questions types in MCScript is shown in Fig.\ref{mcscript_pic_2}. 

%% mcscript的图片一个换成文字版本的table，一个是原图
\begin{figure}
	\begin{framed}
		\begin{tabularx}{\columnwidth}{p{.5cm}XX}
			\textbf{T} & \multicolumn{2}{p{6.5cm}}{I wanted to plant a tree. I went to the home and garden store and picked a nice oak. Afterwards, I planted it in my garden.}\\
			& & \\
			\textbf{Q1} & \multicolumn{2}{l}{What was used to dig the hole?}\\
			& a. a shovel & b. his bare hands\\
			& & \\
			\textbf{Q2} & \multicolumn{2}{l}{When did he plant the tree?}\\
			& a. after watering it & b. after taking it home\\
		\end{tabularx}
	\end{framed}
	\caption{Example questions of MCScript\cite{ostermann2018mcscript}.}
\label{mcscript_pic_1}
\end{figure}

%%%%%%%%ARC

\paragraph{ARC} ARC(AI2 Reasoning Challenge)\cite{clark2018arc} makes use of standardized tests, whose questions are objectively gradable and exhibit the variety in difficulty, which can be a Grand Challenge for AI \cite{clark2018think}\cite{clark2016my}. ARC consists about 7.8K questions. 
\par The authors of ARC also designe two baselines, namely a retrieval-based algorithm and a word co-occurrence algorithm. The Challenge Set, a subset of ARC containing about 2.6K questions, is created by gathering questions that are answered incorrectly by both of these two baselines. The Easy Set is composed of the remaining 5.2K questions. Several state-of-the-art models are tested on the Challenge Set, but none of them are able to significantly outperform a random baseline\cite{clark2018arc}, which reflects the difficulty of the Challenge Set. Two example questions of the Challenge Set questions are as follows: 
\begin{quote}
  {\it Which property of a mineral can be determined just by looking at it? (A) luster {[correct]} (B) mass (C) weight (D) hardness} \\
  \\
 {\it A student riding a bicycle observes that it moves faster on a smooth road than on a rough road. This happens because the smooth road has (A) less gravity (B) more gravity (C) less friction {[correct]} (D) more friction}
\end{quote}
For example, the first question is difficult in that the ground truth, ``Luster can be determined by looking at something'', only appears as a stand-alone sentence in the Web text. However, the incorrect candidate ``hardness'' has a strong correlation with ``mineral'' in the text. 
\par The ARC corpus, a scientific text corpus which  contains 14M science-related sentences and mentions 95\% of the knowledge related to the Challenge Set questions according to a sample analysis \cite{clark2018arc}, is released along with the ARC questions set. The use of the corpus is optional. Some statistics of ARC is shown in Table \ref{stat_arc_1}, Table \ref{stat_arc_2} and Table \ref{stat_arc_1}.

\newcommand\ASC{{\sc ARC}}
\newcommand\Challenge{Challenge}
\newcommand\Additional{Easy}
\newcommand\BUSCSIZE{14M}
\newcommand\ascurl{http://data.allenai.org/arc}
\newcommand\codeurl{https://github.com/allenai/arc-solvers}
\newcommand{\bfit}[1]{\textbf{\textit{#1}}}
\newcommand{\eat}[1]{}

\begin{table}
\setlength{\tabcolsep}{8pt}
\setlength{\doublerulesep}{\arrayrulewidth}
\small
\centering
 \begin{tabular}{|l|rrr|} \hline
%  & \multicolumn{3}{|c|}{{\bf Number of Questions}} \\
 \bigstrut[t] & {\bf \Challenge} & {\bf \Additional} & {\bf \hspace{3ex} Total} \\ \hline \hline
Train \bigstrut[t] & 1119 & 2251 & 3370 \\
Dev & 299 & 570 & 869 \\
Test & 1172 & 2376 & 3548 \\ \hline
\bigstrut[t] TOTAL & 2590 & 5197 & 7787 \\ \hline
 \end{tabular}
 \caption{Number of questions in ARC \cite{clark2018arc} }
 \label{stat_arc_1}
 \end{table}

\begin{table}
\setlength{\tabcolsep}{4pt}
\setlength{\doublerulesep}{\arrayrulewidth}
\small
\centering
\begin{tabular}{|c|rlrl|} \hline
{\bf Grade} \bigstrut[t] & \multicolumn{2}{c}{\bf \Challenge} &	\multicolumn{2}{c|}{\bf \Additional} \\
& \% & (\# qns) & \% & (\# qns) \\
\hline \hline
3 \bigstrut[t] &	3.6 & (94 qns) & 3.4 & (176 qns) \\
4 & 9 & (233) & 11.4  & (591) \\
5 & 19.5 & (506) & 21.2 & (1101) \\
6 & 3.2 & (84) & 3.4 & (179) \\
7 & 14.4 & (372) & 10.7 & (557) \\
8 & 41.4 & (1072) & 41.2 & (2139) \\
9 & 8.8 & (229) & 8.7 & (454) \\ \hline
 \end{tabular}
 \caption{Grade-level distribution of \ASC~questions \cite{clark2018arc}  \label{stat_arc_2}}
 \end{table}

\begin{table}
\setlength{\tabcolsep}{6pt}
\setlength{\doublerulesep}{\arrayrulewidth}
\small
\centering
\begin{tabular}{|l|rr|} \hline
   & \multicolumn{2}{|c|}{\bf min / average / max} \\
  {\bf Property:} \bigstrut[t] & {\bf \Challenge} & {\bf \Additional} \\ \hline
Question (\# words) \bigstrut[t] & 2 / 22.3 / 128 & 3 / 19.4 / 118 \\
Question (\# sentences) & 1 / 1.8 / 11 & 1 / 1.6 / 9 \\
Answer option (\# words) \bigstrut[t] & 1 / 4.9 / 39 & 1 / 3.7 / 26 \\
\# answer options & 3 / 4.0 / 5 & 3 / 4.0 / 5 \\ \hline
 \end{tabular}
 \caption{Properties of the \ASC~Dataset in \cite{clark2018arc} \label{stat_arc_3}}
 \end{table}

% other format
% other format

\paragraph{CoQA} CoQA(Conversational Question Answering systems)\cite{reddy2018coqa} is a conversational style datasets which consists of 126k questions sourced from 8k conversations in 7 different domains. Answers of questions are in free form. The motivation of CoQA is that in daily life human usually get information by asking questions in conversations, and so it is desirable for a machine to be capable of answering such questions. CoQA firstly provides models with a text passage to understand, and then presents a series of questions that appear in a conversation. One example is given in Fig.\ref{coqa_pic_1}.
\par The key challenge of CoQA is that a system must handle conversation history properly to tackle problems like resolving the coreference. Among 7 domains of the passages from which the questions are collected, 2 are used for cross-domain evaluation and 5 are used for in-domain evaluation. The distribution of domains are shown in Table \ref{coqa_pic_2}. Some linguistic phenomena statistics are given in Table \ref{coqa_pic_3}. The coreference and pragmatics are unique and challenging linguistic phenomena that do not appeare in other datasets.

% pic1 of coqa: sample
\begin{figure}
\footnotesize
\centering
\begin{tabular}{p{0.95\columnwidth}}
\midrule
Jessica went to sit in her rocking chair. Today was her birthday and she was turning 80. Her granddaughter Annie was coming over in the afternoon and Jessica was very excited to see her. Her daughter Melanie and Melanie's husband Josh were coming as well. Jessica had $\ldots$\\
\\
Q$_1$:		Who had a birthday? \\
A$_1$: Jessica \\
R$_1$: Jessica went to sit in her rocking chair. Today was her birthday and she was turning 80.\\
\vspace{0em}
Q$_2$: How old would she be?\\
A$_2$: 80 \\
R$_2$: she was turning 80 \\
\vspace{0em}
Q$_3$: Did she plan to have any visitors?\\
A$_3$: Yes \\
R$_3$: Her granddaughter Annie was coming over \\
\vspace{0em}
Q$_4$: How many?\\
A$_4$: Three \\
R$_4$: Her granddaughter Annie was coming over in the afternoon and Jessica was very excited to see her. Her daughter Melanie and Melanie's husband Josh were coming as well. \\
\vspace{0em}
Q$_5$: Who?\\
A$_5$: Annie, Melanie and Josh \\
R$_5$: Her granddaughter Annie was coming over in the afternoon and Jessica was very excited to see her. Her daughter Melanie and Melanie's husband Josh were coming as well.\\
\bottomrule
\end{tabular}
\caption{A conversation example from the CoQA\cite{reddy2018coqa}. Each turn contains a question (Q$_i$), an answer (A$_i$) and a rationale (R$_i$) that supports the answer.}
\label{coqa_pic_1}
\end{figure}

% pic2 of coqa : stat
\begin{table}
\footnotesize
\centering
\setlength{\tabcolsep}{5mm}{
\begin{tabular}{p{1.9cm}rrrr}
\toprule
Domain & \#Passages & \#Q/A & Passage    & \#Turns per \\
 & & pairs & length & passage \\
\midrule
Children's Sto.  & 750 & 10.5k & 211 &  14.0 \\
Literature  & 1,815 & 25.5k & 284  & 15.6 \\
Mid/High Sch. & 1,911 & 28.6k & 306  & 15.0 \\
News & 1,902 & 28.7k & 268 &  15.1 \\
Wikipedia & 1,821 & 28.0k & 245  & 15.4 \\
\midrule
\multicolumn{5}{c}{Out of domain} \\
\midrule
Science & 100 & 1.5k & 251  & 15.3\\
Reddit & 100 & 1.7k & 361 & 16.6 \\
\midrule
Total & 8,399 & 127k  & 271 & 15.2 \\
\bottomrule
\end{tabular}}
\caption{Distribution of domains in CoQA in \cite{reddy2018coqa}.}
\label{coqa_pic_2}
\end{table}

\begin{table}
\centering
\footnotesize
\setlength{\tabcolsep}{5mm}{
\begin{tabular}{lp{5.5cm}c}
\toprule
\bf Phenomenon & \bf Example & \bf Percentage \\
\midrule
\multicolumn{3}{c}{Relationship between a question and its passage} \\
\midrule
Lexical match & Q: Who had to rescue her?& 29.8\% \\
& A: the coast guard \\
& R: Outen was rescued by the coast guard \\
Paraphrasing & Q: Did the wild dog approach? & 43.0\% \\
& A: Yes \\
& R: he drew cautiously closer \\
Pragmatics &  Q:               Is Joey a male or female?  &  27.2\% \\
 & A:  Male \\
& R: it looked like a stick man so she kept \textbf{him}. She named her new noodle friend Joey \\
\midrule
\multicolumn{3}{c}{Relationship between a question and its conversation history} \\
\midrule
No coref. & Q: What is IFL? & 30.5\% \\
Explicit coref. & Q: Who had Bashti forgotten? & 49.7\% \\
& A: the puppy \\
& Q: What was \textbf{his} name? \\
Implicit coref. & Q: When will Sirisena be sworn in? & 19.8\% \\
& A: 6 p.m local time  \\
& Q: \textbf{Where}?\\
\bottomrule
\end{tabular}}
\caption{Linguistic phenomena in CoQA questions given by paper\cite{reddy2018coqa}.}
\label{coqa_pic_3}
\end{table}

%%%%%%%%%%%%%%%%%
%%%% 模型，技术
%%%%%%%%%%%%%%%%%

%\section{Techniques Employed in MRC}
\section {MRC Techniques}
\label{sec:3}
In this section, we will introduce different techniques employed in MRC.
\subsection{Non-Neural Method}
\label{sec:3.1}
Before the neural networks came into fashion, many MRC systems were developed based on different non-neural techniques, which now mostly serve as baselines for comparison. Next, we will introduce the techniques including TF-IDF, sliding window, logistic regression and boosted method.

\paragraph{TF-IDF} The TF-IDF (term frequency-inverse document frequency) technique is widely used in the Information Retrieval area and finds a place in the MRC tasks later. As validated before\cite{clark2016combining}, if candidate answers are presented, retrieval-based models can serve as a strong baseline. This kind of baseline is widely used in multi-document datasets such as WIKIHOP\cite{ref_wikihop}. By solely exploiting lexical correlation between 
the concatenation of a candidate answer and the query and a given document, this kind of algorithm can predict the candidate with the highest similarity score among all documents. Because the inter-document information is usually ignored by TF-IDF, this baseline can not detect how much a question rely on cross-document reasoning.

\paragraph{Sliding Window} The sliding window algorithm is constructed as a baseline in the dataset MCTest\cite{richardson2013mctest}. It predicts an answer based on simple lexical information in a sliding window. Inspired by TF-IDF, this algorithm uses inverse word count as weight of each word, and maximize the bag-of-word similarity between the answer and the sliding window in the given passage. %The detailed procedure is described as below.
%\par 
%For each hypothesis answer $A_i(i=1,2,3,4)$, a set of words in questions Q and $A_i$ are combined as S, whose length is the same as the sliding window. Then it calculates inverse word count for every word w in the given passage P, which is denoted as IC(w). Now we search for the start position j(j=1,2$,\dots,|P|$) of sliding window in P that can maximize the sum up of IC(w) of words in window that also appear in S.
%The pseudo-code is given as below.

\begin{comment}
% 将原来的伪代码图片换成了代码
\begin{algorithm} 
\centering 
  \caption{Sliding Window Algorithm in paper\cite{richardson2013mctest}}  
  \begin{algorithmic}  
    \For{$i=1$ to $4$}  
      \State $S = A_i\cup Q$
      \State $sw_i = \mathop{max}\limits_{j=1...|P|}\sum\limits_{w=1...|S|} \left\{ \begin{array} { c c } { I C \left( P _ { j + w } \right) } & { \text { if } P _ { j + w } \in S } \\ { 0 } & { \text { otherwise } } \end{array} \right.$;  
    \EndFor  \\
    \Return $sw_{1...4}$
  \end{algorithmic}  
\label{slidingwindow_pic_1}  
\end{algorithm}  
\end{comment}

\paragraph{Logistic Regression}
This baseline method is proposed in SQuAD\cite{ref_squad}. 
%It makes use of a large mount of features extracted from answers including lengths, bigram frequencies, word frequencies, span POS tags, lexical features, dependency tree path features etc, and predicts whether a text-span is the answer based on all those information.
It extracts a large mount of features  from the candidates including lengths, bigram frequencies, word frequencies, span POS tags, lexical features, dependency tree path features etc., and predicts whether a text-span is the final answer based on all those information.

\paragraph{Boosting method}
This model is proposed  as a conventional feature-based baseline for CNN/Daily Mail dataset \cite{chen2016thorough}. Since the task can be seen as a ranking problem---making the score of the predicted answer rank top among all the candidates, the authors turn to the implementation of LambdaMART \cite{wu2010adapting} in Ranklib package\footnote{https://sourceforge.net/p/lemur/wiki/RankLib/.}, a highly successful ranking algorithm using forests of boosting decision trees. Through feature engineering, 8 features templates\footnote{the details can be found in the paper} are chosen to form a feature vector which represents a candidate, and the weight vector will be learnt so that the correct answer will be ranked the highest.

\subsection{Neural-Based Method}
With the popularity of neural networks, end-to-end models have produced promising results on some MRC tasks. These models do not need to design complex manually-devised features that traditional approaches relied on, and perform much better than them. Next we will introduce several end-to-end models, mainly in chronological order.

\begin{figure}
\centering
\includegraphics[width=\textwidth]{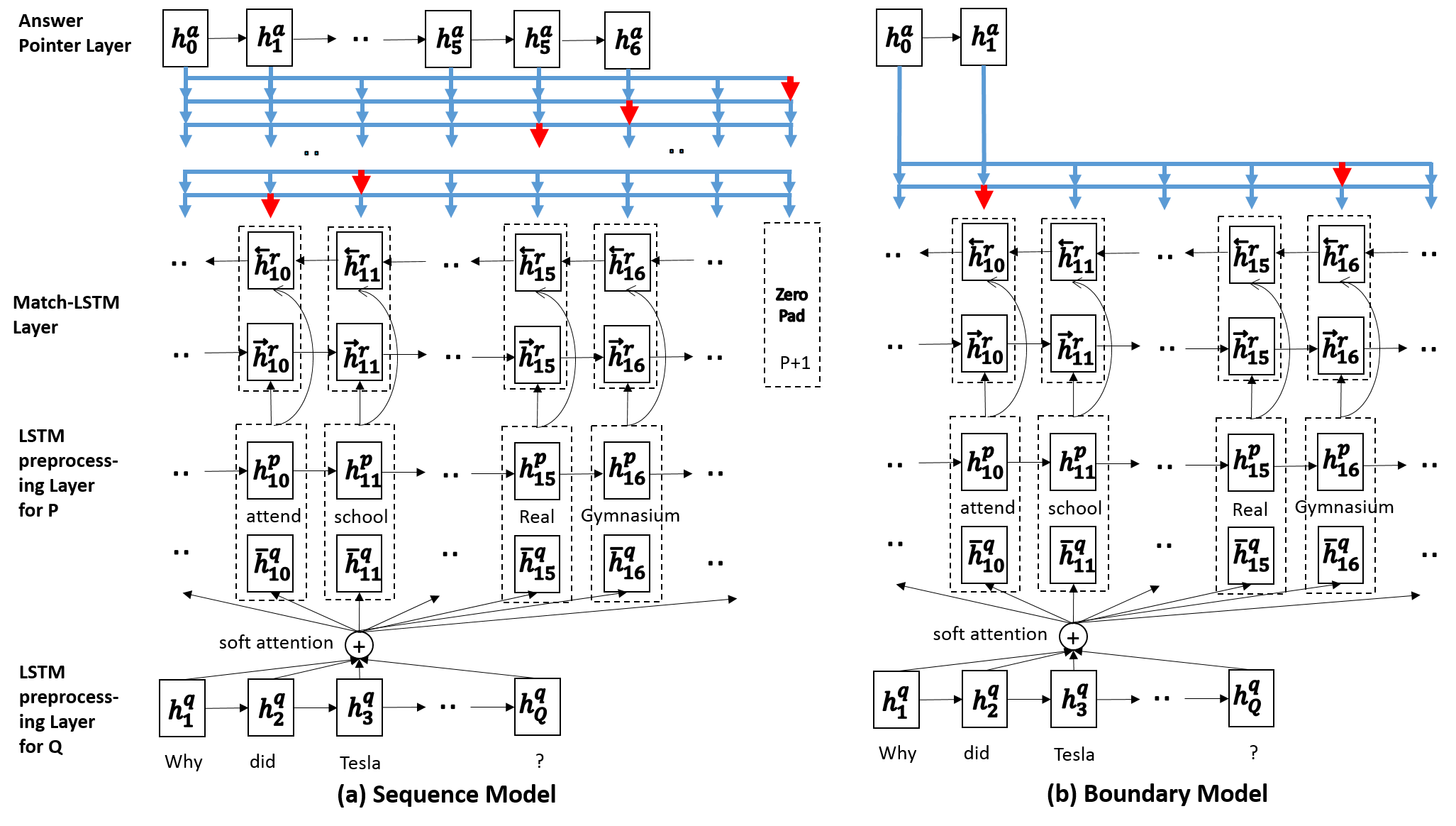}
\caption{the overview of two models in \cite{wang2016machine}}
\label{matchPtr}
\end{figure}

% 是不是第一个存疑一下，是从中文博客上看到的，但是Google没搜索到。。
\paragraph{Match-LSTM+Pointer Network}\label{mlstm+ptr} As the first end-to-end neural architecture\cite{wang2016machine} proposed for SQuAD, this model combines the match-LSTM\cite{wang2015learning}, which is used to get a query-aware representation of passage, and the Pointer Network\cite{2015arXiv150603134V}, which aims to construct an answer so that every token within it comes from the input text. An overall picture of the model architecture is given in Fig.\ref{matchPtr}.
\par Match-LSTM is originally designed for predicting textual entailment. In that task, a premise and a hypothesis are given, and the match-LSTM encodes the hypothesis in a premise-aware way. For every token in hypothesis, this model uses soft-attention mechanism, which will be discussed later in Sect.\ref{sec_attention}, to get a weighted vector representation of premise. This weighted vector is concatenated with a vector representation of the according token, and both are fed into an LSTM, namely the match-LSTM. In this paper, the authors replace the premise and hypothesis with the query and passage to get a query-aware representation of the given passage. Two preprocessing LSTMs are employed respectively to encode the query and the passage. And a bidirectional match-LSTM is employed to obtain the query-aware representation of the passage.
\par After getting the query-aware representation of the passage, a Pointer Network(Ptr-Net) is employed to generate answers by selecting tokens from the input passage. At each inference step, Ptr-Net uses soft-attention mechanism to get a probability distribution of the input sequence, and selects the token with the largest possibility as the output symbol. Moreover, two different strategies are proposed for constructing the answer.
\par The sequence model assumes that every word in the answer can appear in any position in the passage, and the length of the answer is not fixed. In order to tell the model to stop generating tokens after getting the whole answer, a special symbol is placed at the end of the passage, the prediction of this symbol indicates the termination of the answer generating.
\par The boundary model works differently from the Sequence Model in that it only predicts the start indice $a_s$ and the end indice $a_e$, in other word, it's based on the assumption that the answer appears as a continuous segment of the passage. The test result shows an advantage of the boundary model over the other one.

% bidaf的论文中，对结构的描述没什么改写的空间。。。各个Layer的功能描述的改写也比较有限，感觉这部分的重复率会挺高 T_T
\paragraph{Bi-Directional Attention Flow} Proposed by \cite{seo2016bidirectional}, the Bi-Directional Attention Flow has two key features at the context encoding stage. First, this model takes different levels of granularity as input, including character-level, word-level and contextualized embeddings. Second, it uses bi-directional attention flow, namely a passage-to-query attention and a query-to-passage attention, to get a query-aware passage representation. The detailed description is given as follows.
\par
\begin{figure}
\centering
\includegraphics[width=\textwidth]{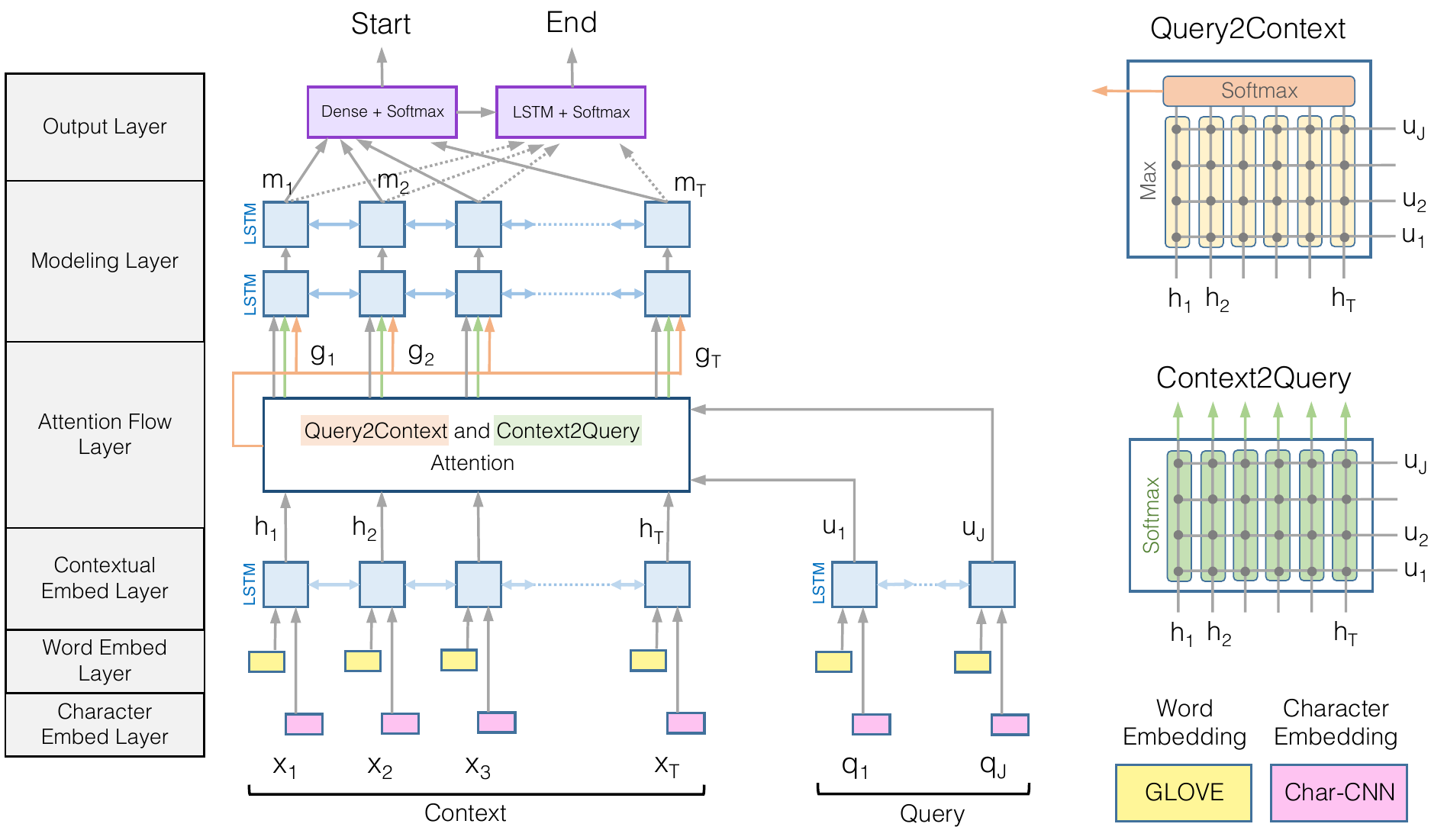}
\caption{Overview of BiDAF architecture given in \cite{seo2016bidirectional}.}
\label{bidaf_pic_1}
\end{figure}

 As is shown in Fig.\ref{bidaf_pic_1}, the BiDAF model has six layers. The \textbf{Character Embedding layer} and the \textbf{Word Embedding Layer} map each each word into the vector space based respectively on character-level CNNs\cite{kim2014convolutional} and the pre-trained GloVe embedding\cite{pennington2014glove}. The concatenation of these two word embeddings is passed to a two layer Highway Networks\cite{srivastava2015highway}, the output of which is provided to a bi-directional LSTM in the \textbf{Contextual Embedding Layer} to refine the word embedding using the context information. These first three layers are applied to both the query and the passage.
\par The \textbf{Attention Flow Layer} is where the information from the query and the passage mixed and interacted. Instead of summarizing the passage and the query into a fixed vector like most attention mechanisms do, this layer grants raw information including attention vectors and the embeddings from previous layers flowing to the subsequent layer, which reduces the information loss. The attentions are computed in two directions---from passage to query and from query to passage. The detailed information of the  \textbf{Attention Flow Layer} will be given in Sect.\ref{sec_attention}.

\par The\textbf{ Modeling Layer} takes in the query-aware representation of context words and used two bi-directional LSTM to capture the interactions among the passage words according to the query. The last \textbf{Output Layer} is task-specific, which gives the prediction of the answer.

\paragraph{Gated Attention}Gated-Attention Reader\cite{dhingra2016gated} targets at realizing multi-hop reasoning in answering cloze-style questions over documents. A multiplicative interaction between the query and the hidden state of the document is employed in its attention mechanism. The multi-hop architecture of the model imitates the multi-step reasoning of human in reading comprehension.
\par
\begin{figure}
\centering
\includegraphics[width=\textwidth]{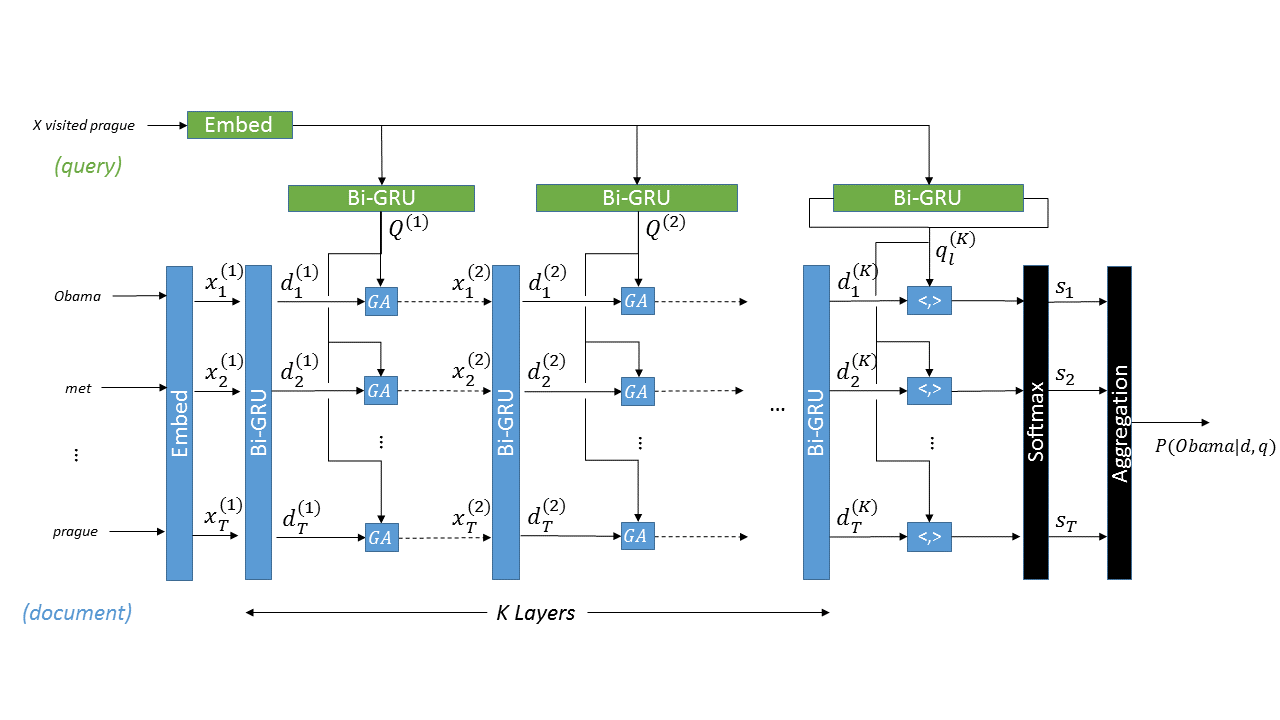}
\caption{Gated Attention architecture given in \cite{dhingra2016gated}.}
\label{ga_pic_1}
\end{figure}

 The overview of the model is given in Fig.\ref{ga_pic_1}. The model reads the document and the query iteratively in a row of K layers. In $k$th layer, first, the model uses bidirectional Gated Recurrent Unit(Bi-GRU)\cite{cho2014learning} to transform the $X^{(k-1)}$, embeddings of document passed from the last layer, to get $D^{(k)}$. Then a layer-specific query representation is transformed by another Bi-GRU to get $Q^{(k)}$. 
$$
\begin{aligned}
D ^ { ( k ) } &= \stackrel { \longleftrightarrow  } { \mathrm { GRU } } _ { D } ^ { ( k ) } \left( X ^ { ( k - 1 ) } \right)\\
Q ^ { ( k ) } &= \stackrel { \longleftrightarrow  } { \mathrm { GRU } } _ { Q } ^ { ( k ) } \left( Y \right) \\
\end{aligned}
$$
Then both $D^{(k)}$ and $Q^{(k)}$ are fed to a \textit{Gated Attention} module, the result of which, $X^{(k)}$, will be passed to the next layer.
\par For each token $d_i$ in $D ^ { ( k ) }$, the \textit{Gated Attention} module uses soft attention to get a token specified representation of query: $\tilde { q }_i$. Finally we get the new embeddings of this token, $x _ { i }$, by applying a element-wise multiplication for $\tilde { q }_i$ and $d_i$.
$$
\begin{aligned} \alpha _ { i } & = \operatorname { softmax } \left( Q ^ { \top } d _ { i } \right) \\
\tilde { q } _ { i } & = Q \alpha _ { i } \\ x _ { i } & = d _ { i } \odot \tilde { q } _ { i } \end{aligned}
$$
\par At the last stage, the decoder employs a softmax layer to the inner-product between outputs of last layer to get the possibility distribution of the predict answers.

\paragraph{DCN} Dynamic Coattention Networks(DCN)\cite{xiong2016dynamic} introduces coattention mechanism to combine co-dependent representations of query and the document, and dynamic iteration to avoid been trapped in local maxima corresponding to incorrect answers like previous single-pass models. The Dynamic pointer decoder takes in the output of coattention encoder and generates the final predictions. Detailed procedures is given as follows.

\par Let $\left(x_{ 1 } ^ { Q } , x _ { 2 } ^ { Q } , \ldots , x _ { n } ^ { Q } \right)$ denote the sequence of embeddings of words in query and $\left( x _ { 1 } ^ { D } , x _ { 2 } ^ { D } , \ldots , x _ { m } ^ { D } \right)$ for those in document. The the details of DCN are as follows.
\par In the Document and Question encoder, the vector representations of the document and the query are fed into LSTM respectively, and the hidden states at each step are combined to form the encoding matrix $D = \left[ d _ { 1 } \ldots d _ { m } d _ { \varnothing } \right] \in \mathbb { R } ^ { \ell \times ( m + 1 ) }$ and $Q ^ { \prime } = \left[ q _ { 1 } \quad \ldots q _ { n } q _ { \varnothing } \right] \in \mathbb { R } ^ { \ell \times ( n + 1 ) }$. Sentinel vector $d _ { \varnothing }$ and $q _ { \varnothing }$ \cite{merity2016pointer} is appended to the encoding matrix to enable the model to map some unrelated words that exclusively appear in either the query or the document to this void vector. To allow for some variation between the document encoding space and the query encoding space, a non-linear projection $Q = \tanh \left( W ^ { ( Q ) } Q ^ { \prime } + b ^ { ( Q ) } \right) \in \mathbb { R } ^ { \ell \times ( n + 1 ) }$is applied to $Q'$. The final representations of the document and the query are $D$ and $Q$.
\par The Coattention encoder takes in $D$ and $Q$ and outputs coattention encoding matrix $U =\left[ u _ { 1 } , \dots , u _ { m } \right] \in \mathbb { R } ^ { 2 \ell \times m }$, which is the input to the Dynamic pointing decoder. The details of Coattention encoder will be discussed in Sec.\ref{sec_attention}.
\par
\begin{figure}
\centering
\includegraphics[width=\textwidth]{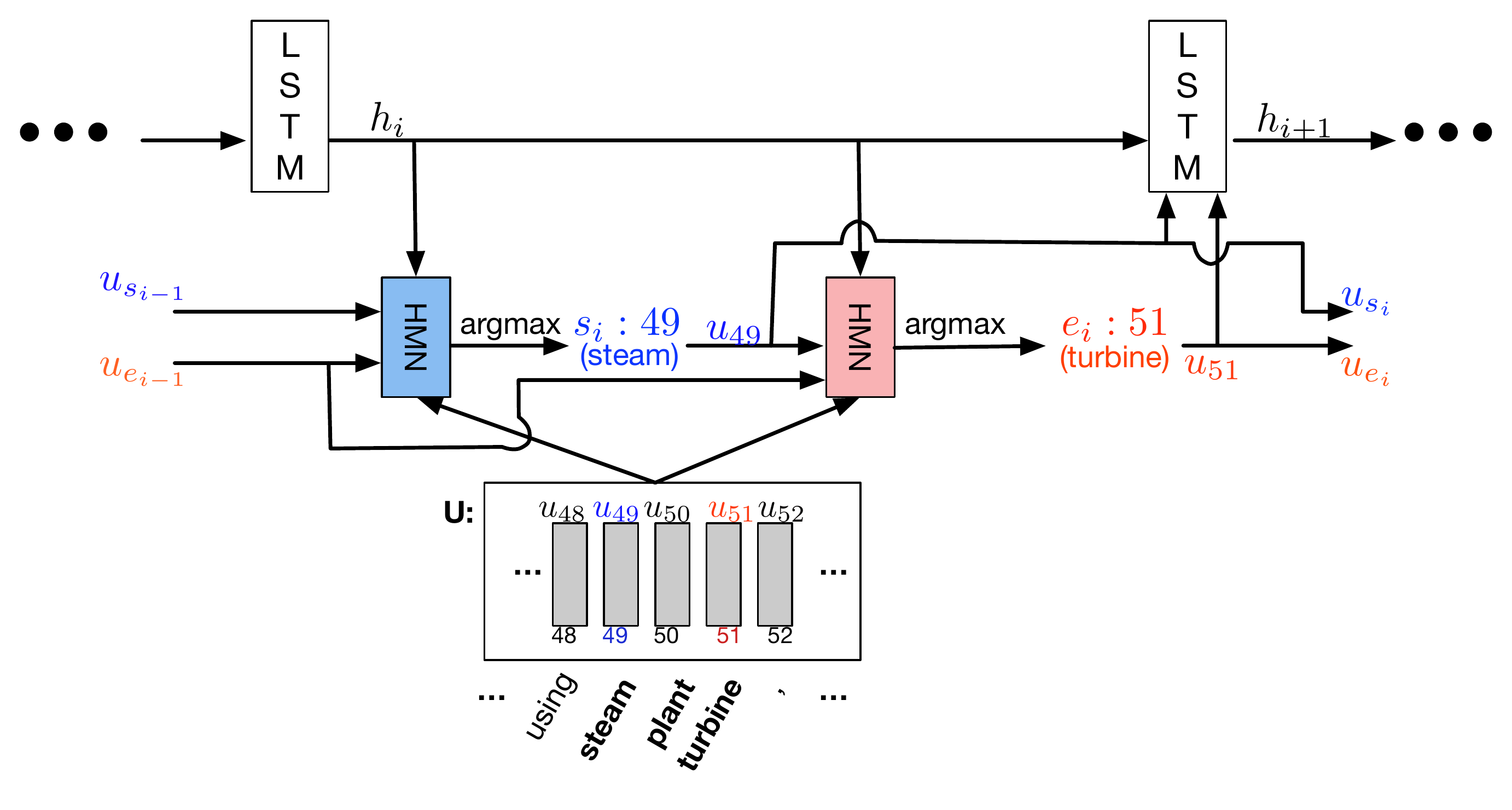}
\caption{Architecture of Dynamic Decoder from paper\cite{xiong2016dynamic}. Blue denotes the variables and functions related to estimating the start position whereas red denotes the variables and functions related to estimating the end position.}
\label{DCN_2}
\end{figure}

 The overview of Dynamic pointing decoder is given in Fig.\ref{DCN_2}. To enable the model to recover from local maxima, the Highway Maxout Network (HMN) is proposed to predict the start point and the end point iteratively. HMN is composed of Highway Networks\cite{srivastava2015highway}, which is characterized by the skip connect that passes gradient effectively through deep networks, and Maxout Networks\cite{goodfellow2013maxout}, a learnable activation function that has strong empirical performance.
\par During the iteration, the hidden state of the decoder is updated according to Eq.1.
$$
h _ { i } = \mathop{ LSTM } _ { d e c } \left( h _ { i - 1 } , \left[ u _ { s _ { i - 1 } } ; u _ { e _ { i - 1 } } \right] \right) \eqno{(1)}
$$
where $u _ { s_{ i-1}}$ and $u _ { e_{ i-1}}$ are the coattention representations of according start and end words predicted by (i-1)th iteration. Given $h_i$, $u _ { s_{ i-1}}$ and $u _ { e_{ i-1}}$, the possibility of the \textit{t}th word to be the start or the end point is calculated by Eq.2.
$$
\alpha _ { t } = \mathrm { HMN }  \left( u _ { t } , h _ { i } , u _ { s _ { i - 1 } } , u _ { e _ { i - 1 } } \right)\eqno{(2)}
$$
the word with the maximum possibility is selected as the prediction at current step.
\par
\begin{figure}
\centering
\includegraphics[width=0.5\textwidth]{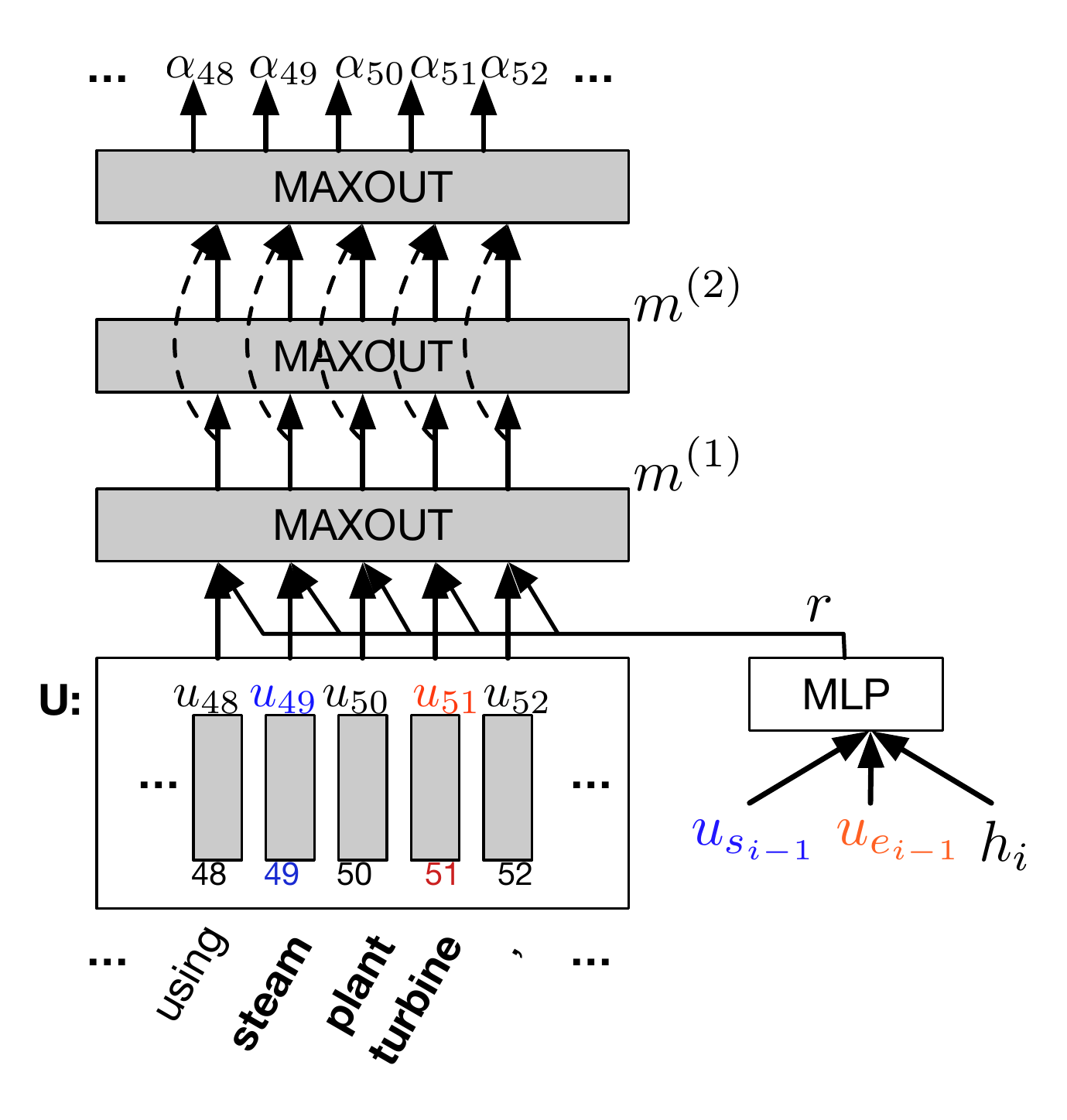}
\caption{Architecture of Highway Maxout Network given in \cite{xiong2016dynamic}.}
\label{DCN_3}
\end{figure}

 The architecture of HMN is given in Fig.\ref{DCN_3}. The mathematical description of HMN is given as follows:

$$
\begin{aligned} \operatorname { HMN } \left( u _ { t } , h _ { i } , u _ { s _ { i - 1 } } , u _ { e _ { i - 1 } } \right) & = \max \left( W ^ { ( 3 ) } \left[ m _ { t } ^ { ( 1 ) } ; m _ { t } ^ { ( 2 ) } \right] + b ^ { ( 3 ) } \right) \\ r & = \tanh \left( W ^ { ( D ) } \left[ h _ { i } ; u _ { s _ { i } - 1 } ; u _ { e _ { i - 1 } } \right] \right) \\ m _ { t } ^ { ( 1 ) } & = \max \left( W ^ { ( 1 ) } \left[ u _ { t } ; r \right] + b ^ { ( 1 ) } \right) \\ m _ { t } ^ { ( 2 ) } & = \max \left( W ^ { ( 2 ) } m _ { t } ^ { ( 1 ) } + b ^ { ( 2 ) } \right) \end{aligned}
$$
where r is a non-linear projection of the current state.

% 删除换成文字
% \begin{figure}
% \centering
% \includegraphics[width=\textwidth]{DCN_1.pdf}
% \caption{\cite{xiong2016dynamic}}
% \label{DCN_1}
% \end{figure}
%

\begin{figure}
\centering
\includegraphics[width=1\textwidth]{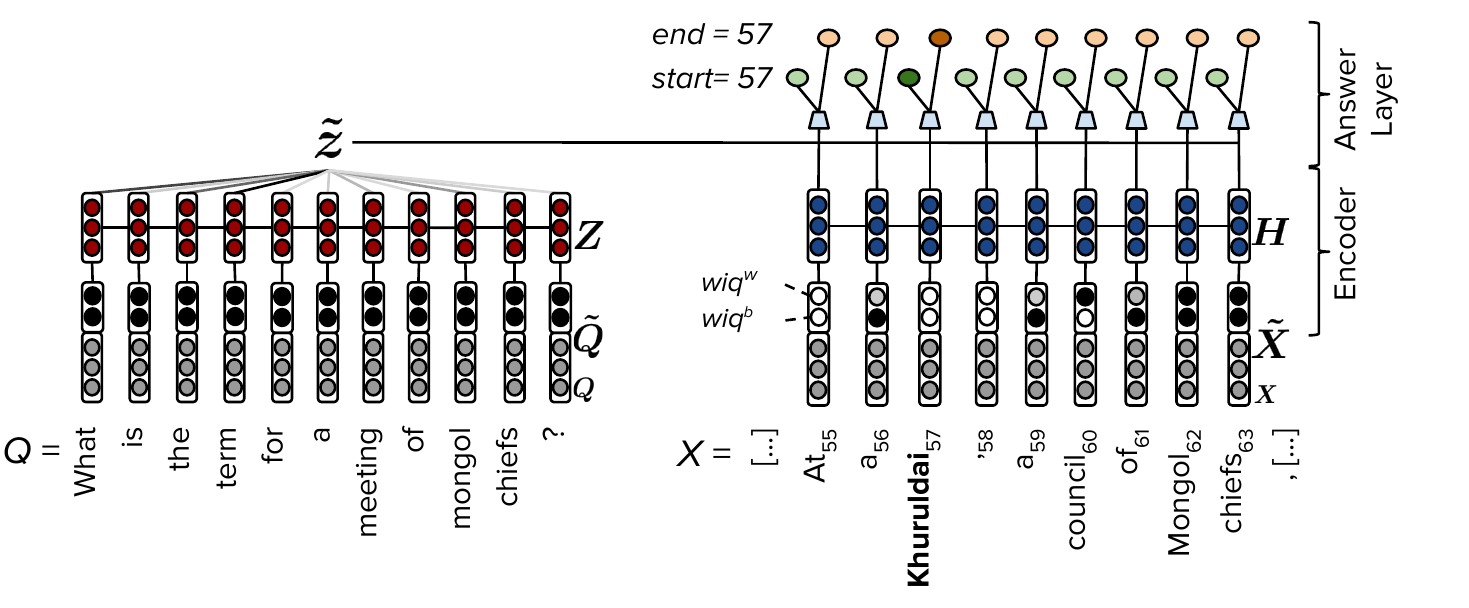}
\caption{Overview of FastQA architecture from \cite{WeissenbornWS17}.}
\label{fastqa_pic_1}
\end{figure}

\paragraph{FastQA} FastQA\cite{WeissenbornWS17} achieved competitive performance with simple architecture, which questions the necessity of improving complexity of QA systems. Unlike many systems that employed a complex interaction layer to catch the interaction between the query and the context, FastQA only makes use of computable features on word levels. The overview of FastQA architecture is given in Fig.\ref{fastqa_pic_1}.
\par The binary word-in-question($wiq^b$) feature indicates whether a token in passages appears in the corresponding query.
$$
\mathrm { wiq } _ { j } ^ { b } = \mathbb { I } \left( \exists i : x _ { j } = q _ { i } \right)
$$
\par The weighted feature($wiq^w$) which is defined as below takes the term-frequency and the similarity between query and context into account.
$$
\begin{aligned} \operatorname { sim } _ { i , j } & = \boldsymbol { v } _ { w i q } \left( \boldsymbol { x } _ { j } \odot \boldsymbol { q } _ { i } \right) \quad , \boldsymbol { v } _ { w i q } \in \mathbb { R } ^ { n } \\ \operatorname { wiq } _ { j } ^ { w } & = \sum _ { i } \operatorname { softmax } \left( \operatorname { sim } _ { i , } \right) _ { j } \end{aligned}
$$
\par The concatenation of these two features and the original representation of each words is fed into a Bi-LSTM to get the final hidden state. The Answer Layer is composed of a simple 2-layer feed-forward network along with a beam search.

\begin{figure}
\centering
\includegraphics[width=1\textwidth]{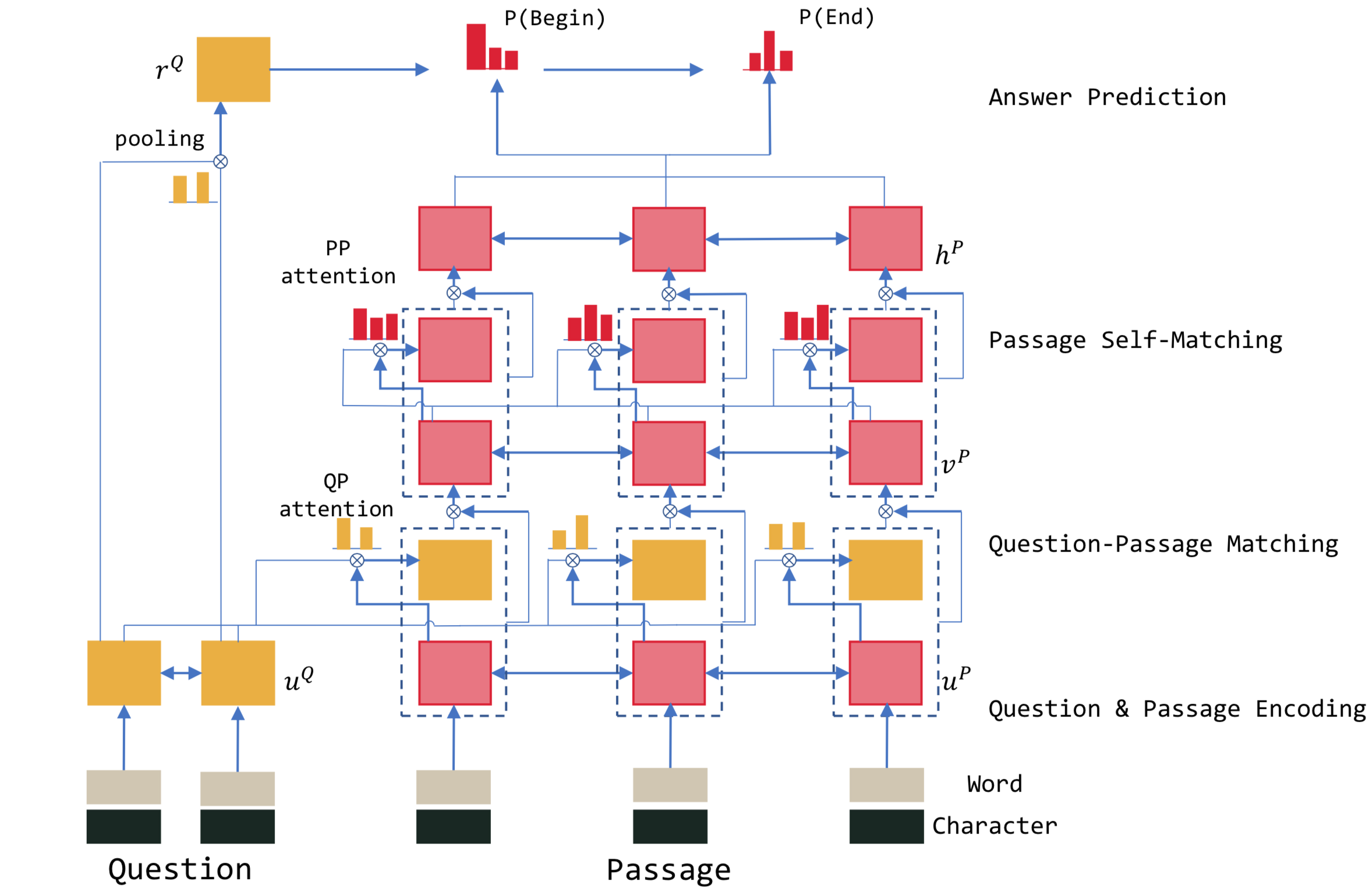}
\caption{Overview of the R-net architecture from paper\cite{wang2017gated}}
\label{rnet_pic_1}
\end{figure}

\paragraph{R-NET} \label{ref_rnet}The R-NET\cite{wang2017gated} was proposed in 2017 by MSRA and achieved state-of-the-art results on SQuAD and MS-MARCO. An overview of its architecture is shown in Fig.\ref{rnet_pic_1}.
\par Given the word-level and character-level embeddings, R-NET firstly employs a bi-directional GRU\cite{cho2014learning} to encode the questions and passages. Then it uses a gated attention-based recurrent network to fuse the information from the question and passage. Later a self-matching layer is used to fine-tune and get the final representation of the passage. The output layer is based on pointer networks similar to that in match-LSTM to predict the boundary of the answer. The initial hidden vectors of the pointer network are computed by an attention-pooling over the final passage representations.
\par The gated attention-based recurrent network adds another gate to normal attention-based recurrent networks. This gate gives the weight of certain passage information according to the question. Inspired by \cite{rocktaschel2015reasoning},  the sentence-pair representations are obtained $\left\{ v _ { t } ^ { P } \right\} _ { t = 1 } ^ { n }$ as follows:
$$
\begin{aligned}
v _ { t } ^ { P } &= \operatorname { RNN } \left( v _ { t - 1 } ^ { P } , \left[ u _ { t } ^ { P } , c _ { t } \right]^* \right) \\  
\left[ u _ { t } ^ { P } , c _ { t } \right] ^ { * } &= g _ { t } \odot \left[ u _ { t } ^ { P } , c _ { t } \right]  \\  
 s _ { j } ^ { t } &= \mathrm { v } ^ { \mathrm { T } } \tanh \left( W _ { u } ^ { Q } u _ { j } ^ { Q } + W _ { u } ^ { P } u _ { t } ^ { P } + W _ { v } ^ { P } v _ { t - 1 } ^ { P } \right) \\    a _ { i } ^ { t } &= \exp \left( s _ { i } ^ { t } \right) / \Sigma _ { j = 1 } ^ { m } \exp \left( s _ { j } ^ { t } \right)  \\   
 c _ { t } &= \Sigma _ { i = 1 } ^ { m } a _ { i } ^ { t } u _ { i } ^ { Q } 
\end{aligned}
$$
where $ g _ { t } = \operatorname { sigmoid } \left( W _ { g } \left[ u _ { t } ^ { P } , c _ { t } \right] \right) $ is the added gate, $\left\{ u _ { t } ^ { P } \right\} _ { t = 1 } ^ { n }$ and $\left\{ u _ { t } ^ { Q } \right\} _ { t = 1 } ^ { m }$ are original representations of the passage and the question.
\par To exploit information from the whole passage for each token, a self-matching attention is applied to get the final representation of the passage $h^P$. The details of self-matching attention is given in Sec.\ref{sec_attention}. 
\par The Output Layer uses pointer networks\cite{vinyals2015pointer} to predict the start and end position of the answer. The initial hidden vector for the pointer network is an attention-pooling over the question representation $h^P$. The objective function is the sum of the negative log probabilities of the ground truth start and end position by the predicted distributions.

%% R-NET 的out-put再加一点吧

%

\begin{figure}
\centering
\includegraphics[width=1\textwidth]{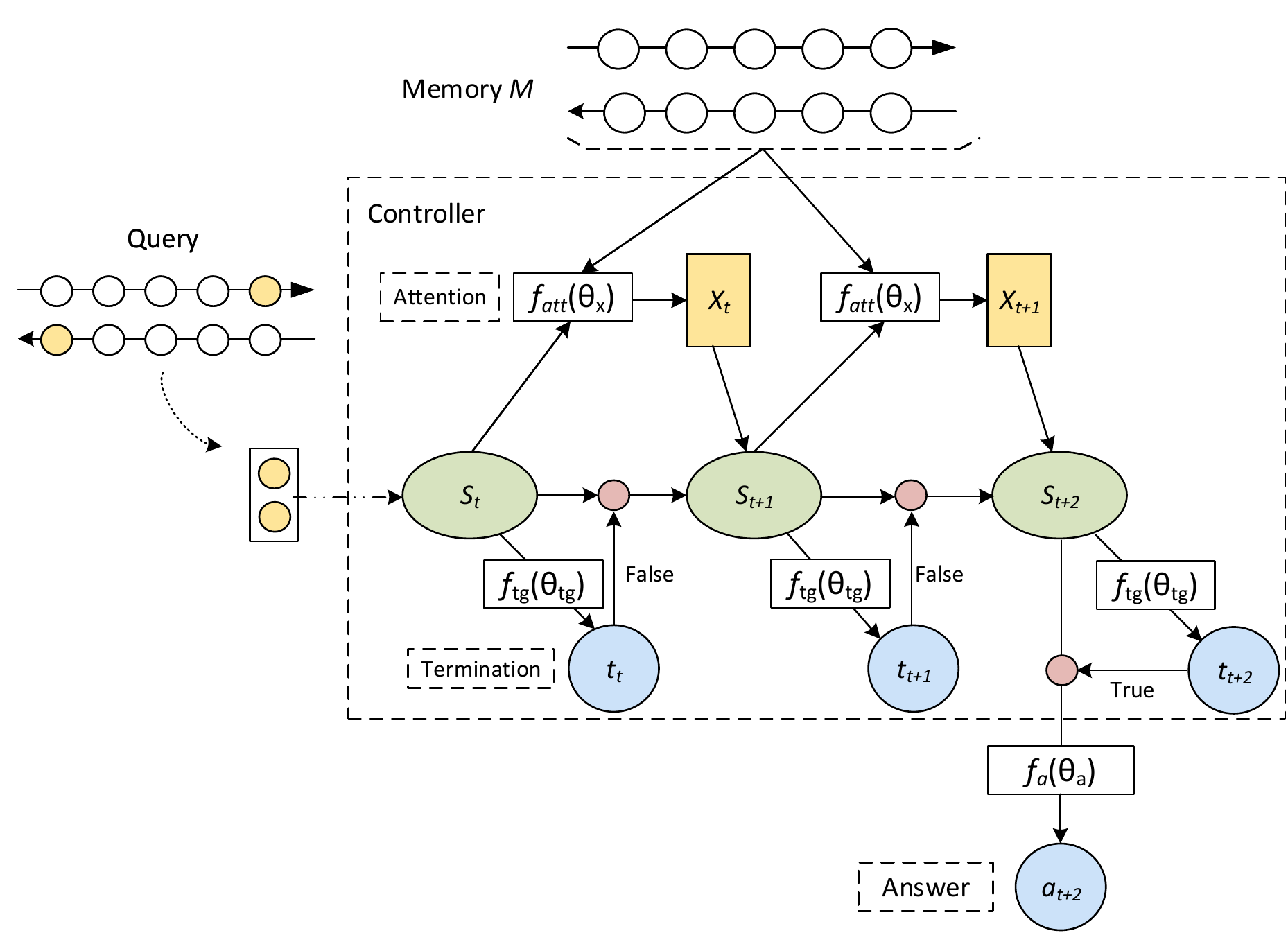}
\caption{Overview of ReasoNet structure  from \cite{shen2017reasonet}.}
\label{reaso_pic_1}
\end{figure}

\paragraph{ReasoNet} Unlike previous models which have fixed number of turns during reading or reasoning regardless of the complexity of queries and passages, the ReasoNet\cite{shen2017reasonet} makes use of reinforcement learning to dynamically determine the reading and reasoning depth. The intuition of this work comes from that the difficulty of different questions can vary a lot in the same dataset\cite{chen2016thorough}, and the fact that human usually revisit important part of passage and question to answer the question better. An overview of ReasoNet structure is given in Fig.\ref{reaso_pic_1}.
\par The external \textbf{Memory} M is usually the word embeddings encoded by a Bi-RNN. The \textbf{Internal State} $ s $ is updated according to $ s _ { t + 1 } = \operatorname { RNN } \left( s _ { t } , x _ { t } ; \theta _ { s } \right) $, where $ x_t $ is the \textbf{Attention} vector : $ x _ { t } = f _ { a t t } \left( s _ { t } , M ; \theta _ { x } \right) $. The \textbf{Termination Gate} determines when to stop updating states above and predict the answers according to the binary variable $ t_t $: $t _ { t } \sim p ( \cdot | f _ { t g } \left( s _ { t } ; \theta _ { t g } \right) ) )$. In this way, the ReasoNet can mimic the inference process of human, exploit the passages and answer the questions better.

\begin{figure}
\centering
\includegraphics[width=1\textwidth]{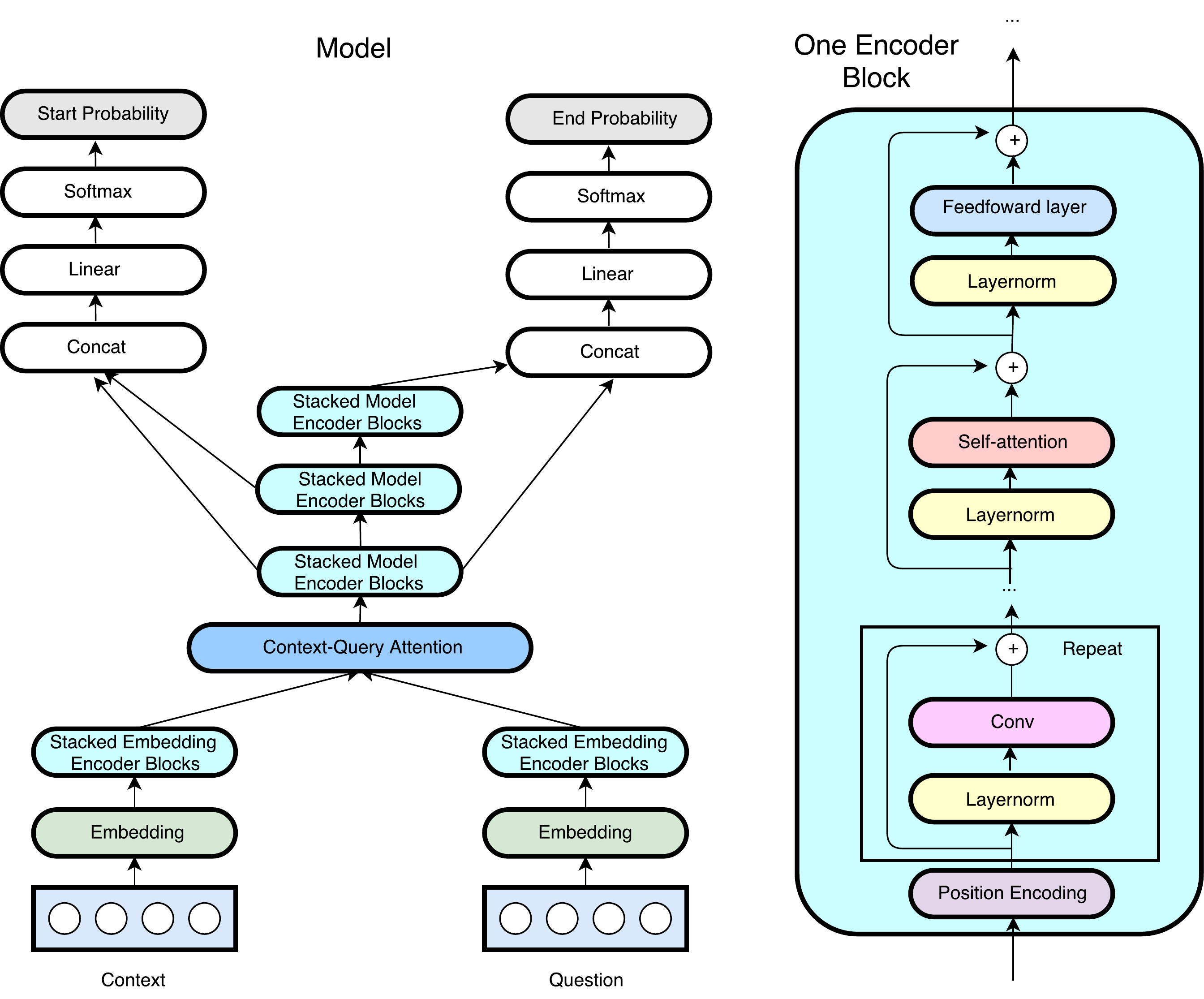}
\caption{Overview of the QANet architecture (left) which has several Encoder Blocks. All Encoder Blocks are the same except that the number of convolutional layers for each block(right) varies. From \cite{yu2018qanet}.}
\label{qanet_pic_1}
\end{figure}

\paragraph{QAnet} Most of the models above are primarily based on RNNs with attention, therefore are often slow for both training and inference due to the sequential nature of RNNs. To make machine comprehension fast, the QAnet\cite{yu2018qanet} are proposed without RNNs in its architecture. An overview of QAnet structure is given in Fig.\ref{qanet_pic_1}.
\par The key difference between QAnet and the previous models is that, QAnet only use convolutional and self-attention mechanism in its embedding and modeling encoders, discarding the commonly used RNNs. The depthwise separable convolutions\cite{chollet2017xception}\cite{kaiser2017depthwise} can capture the local structure of the text, and the multi-head (self-)attention mechanism\cite{vaswani2017attention} will model global interactions within the whole passages. A query-to-context attention similar to that in DCN\cite{xiong2016dynamic} is applied afterwards.
\par The QAnet achieved state-of-the-art accuracy while achieving up to 13x speedup in training and 9x per training iteration, compared to the RNN counterparts\cite{yu2018qanet}.

\subsection{Attention}
\label{sec_attention}
\par The Attention mechanisms have shown great power in selecting important information, aligning and capturing similarity between different part of input. Next we will introduce several representative attention mechanism primarily based on time order.

\paragraph{Hard Attention} was proposed in image caption task in \cite{xu2015show} as the "stochastic hard attention". Let $a = \left\{ \mathbf { a } _ { 1 } , \ldots , \mathbf { a } _ { L } \right\} , \mathbf { a } _ { i } \in \mathbb { R } ^ { D }$ denote the feature vectors captured by CNN, each corresponding to a part of the image. When deciding which one of all features is to feed to the decoder LSTM to generate caption, a one-hot variable $s_{t,i}$ is defined. The indicator $s_{t,i}$ is set to 1 if the $i$-th vector of $a$ is the one used to extract visual features at current step $t$. If we denote the input of decoder LSTM as $ \hat { \mathbf { z } } _ { t } $:
$$
\hat { \mathbf { z } } _ { t } = \sum _ { i } s _ { t , i } \mathbf { a } _ { i }
$$
The paper assigns a multinoulli distribution parametrized by ${α_{t,i}}$ and view $\hat { \mathbf { z } } _ { t }$ as a random variable:
$$
\begin{array}{c}
{p \left( s _ { t , i } = 1 | s _ { j < t } , \mathbf { a } \right) = \alpha _ { t , i }}\\ 
{e _ { t i }  = f _ {  att } \left( \mathbf { a } _ { i } , \mathbf { h } _ { t - 1 } \right)} \\ 
{\alpha _ { t i } = \frac { \exp \left( e _ { t i } \right) } { \sum _ { k = 1 } ^ { L } \exp \left( e _ { t k } \right) }}
\end{array}
$$
where $f _ {  att }$ is a multilayer perceptron. After defining the objective function $L_s$ as below:
$$
\begin{aligned} L _ { s } & = \sum _ { s } p ( s | \mathbf { a } ) \log p ( \mathbf { y } | s , \mathbf { a } ) \\ & \leq \log \sum _ { s } p ( s | \mathbf { a } ) p ( \mathbf { y } | s , \mathbf { a } ) \\ & = \log p ( \mathbf { y } | \mathbf { a } ) \end{aligned}
$$
and approximate its gradient by a Monte Carlo method, the final learning rule for the model is then:
$$
\frac { \partial L _ { s } } { \partial W } \approx \frac { 1 } { N } \sum _ { n = 1 } ^ { N } \left[ \frac { \partial \log p ( \mathbf { y } | \tilde { s } ^ { n } , \mathbf { a } ) } { \partial W } + \right.  \\ \lambda _ { r } ( \log p ( \mathbf { y } | \tilde { s } ^ { n } , \mathbf { a } ) - b ) \frac { \partial \log p \left( \tilde { s } ^ { n } | \mathbf { a } \right) } { \partial W } + \lambda _ { e } \frac { \partial H \left[ \tilde { s } ^ { n } \right] } { \partial W } ]
$$
where the $\lambda _ { r }$ and $\lambda _ { e }$ are two hyperparameters set by crossvalidation.
\par Although hard attention is tricky and troublesome in training, once trained well, it can perform better than soft attention for the sharp focus on memory provided (\cite{Shankar2018SurprisinglyEH}\cite{xu2015show}\cite{ShankarS17}).

\paragraph{Soft Attention} Here we will first introduce the basic form of soft attention in Neural Machine translation task, then we will talk about its variants in other tasks like natural language inference(NLI) and MRC. 
\par Different to hard attention, soft attention calculates a weight distribution among all the input representations, and use the weighted sum of them as the input to the decoder. For example, in \cite{BahdanauCB14}, let $\left( h _ { 1 } , \cdots , h _ { T _ { x } } \right)$ denote the Encoder's output sequence, and $\alpha_{ij}$ denote the weight of each $h_j$ (which indicates to what extent is $h_j$ related to the current output token $t_i$). Then the input to the decoder $c_i$ is :
$$
c _ { i } = \sum _ { j = 1 } ^ { T _ { x } } \alpha _ { i j } h _ { j }
$$
The weights are calculated and learn through a feedforward neural network $a$. 
$$
\alpha _ { i j } =  \frac { \exp \left( e _ { i j } \right) } { \sum _ { k = 1 } ^ { T _ { x } } \exp \left( e _ { i k } \right) } \\ e _ { i j } = a \left( s _ { i - 1 } , h _ { j } \right)
$$
\par In NLI task, input has two components, namely a premise and a hypothesis. And attention is used to exploit the interaction/relation between these two parts. Take the match-LSTM\cite{wang2015learning} as example, we denote $\mathbf { h } _ { j } ^ { s } ​$ and $ \mathbf { h } _ { k } ^ { t }​$ as the resulting hidden states of the Encoder LSTM separately for premise and hypothesis. When predicting the label of the hypothesis, an attention-weighted combinations of the hidden states of the premise is computed through a match-LSTM:
$$
\begin{array}{l}
{\mathbf { a } _ { k } = \sum _ { j = 1 } ^ { M } \alpha _ { k j } \mathbf { h } _ { j } ^ { s }}\\
{\alpha _ { k j } = \frac { \exp \left( e _ { k j } \right) } { \sum _ { j ^ { \prime } } \exp \left( e _ { k j ^ { \prime } } \right) }}\\
{e _ { k j } = \mathbf { w } ^ { \mathrm { e } } \cdot \tanh \left( \mathbf { W } ^ { s } \mathbf { h } _ { j } ^ { s } + \mathbf { W } ^ { t } \mathbf { h } _ { k } ^ { t } + \mathbf { W } ^ { \mathrm { m } } \mathbf { h } _ { k - 1 } ^ { \mathrm { m } } \right)}
\end{array}
$$
where $\mathbf { a } _ { k }$ is the \textit{attention vector} stated above, $\mathbf { w } ^ { \mathrm { e } }$ $\mathbf { W } ^ { s }$ $\mathbf { W } ^ { t }$ $\mathbf { W } ^ { m }$ is the parameters to be learned, and $\mathbf { h } _ { k - 1 } ^ { \mathrm { m }}$ is the hidden state  of match-LSTM at position $k-1$. Finally $\mathbf { a } _ { k }$ is concatenated with $\mathbf { h } _ { k } ^ { t }$ for predicting the result.
\par In MRC task, we can regard the question as a premise and the passage as a hypothesis, as it likes in the model Match-LSTM+Pointer Network. By applying the attention mechanism, we can get additional query information for each token in the passage, which will improve the model performance.

\par Compared to hard attention, soft attention's advantage is that it is differentiabile thus easy to train, and fast in training and inference.

\paragraph{Bi-directional Attention} was proposed in BiDAF. Compared to the attention described above, it considers attention in two directions, or Query-to-context(Q2C) Attention and Context-to-query(C2Q) Attention. Take BiDAF as example, given $\mathbf { H }$ and $\mathbf { U }$, the concatenation of the outputs of the LSTMs in \textbf{Contextual Embedding Layer}, the similarity matrix $\mathbf { S }$ is computed:
$$
\begin{array}{c}
{\mathbf { S } _ { t j } = \alpha \left( \mathbf { H } _ { : } , \mathbf { U } _ { : j } \right)}\\
{\alpha ( \mathbf { h } , \mathbf { u } ) = \mathbf { w } _ { ( \mathbf { S } ) } ^ { \top } [ \mathbf { h } ; \mathbf { u } ; \mathbf { h } \circ \mathbf { u }]}
\end{array}
$$
where $\mathbf { w } _ { ( \mathbf { S } ) } ^ { \top }$ is trainable parameters, $\circ$ is elementwise multiplication. Then we can compute the C2Q attention weights and the attended query vectors by:
$$
\begin{array}{c}
{\mathbf { a } _ { t } = \operatorname { softmax } \left( \mathbf { S } _ { t : } \right)} \\  
{\tilde { \mathbf { U } } _ { : t } = \sum _ { j } \mathbf { a } _ { t j } \mathbf { U } _ { : j }}
\end{array}
$$
Similarily the Q2C attention weights and attended context vectors are:
$$
\begin{array}{c}
{\mathbf { b } = \operatorname { softmax } \left( \max _ { c o l } ( \mathbf { S } ) \right)}\\
{\tilde { \mathbf { h } } = \sum _ { t } \mathbf { b } _ { t } \mathbf { H } _ { : t }}
\end{array}
$$
Finally two attention vectors above are combined together with the original contextual embeddings $\mathbf{H}$ through a vector fusing function, the result of which serve as the base for future modeling or prediction.
\par The Bi-directional Attention adds more information in the Q2C Attention part compared to normal attention mechanism. However, as shown in the ablation study of \cite{seo2016bidirectional}, the attention in this direction is less useful than the standard C2Q Attention(on SQuAD dev set). The reason is that the query is usually short, and the added Q2C information is relatively small than that of the other one.

\begin{figure}
\centering
\includegraphics[width=\textwidth]{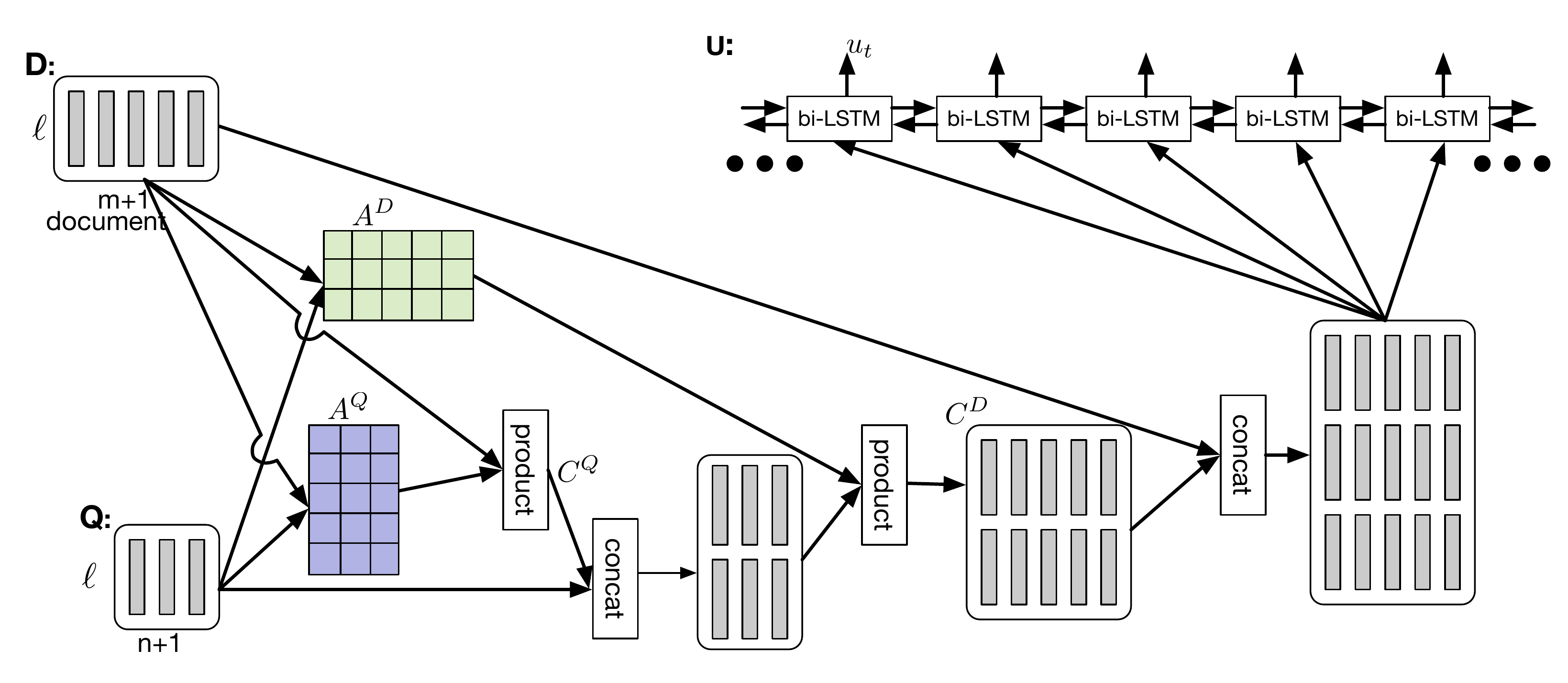}
\caption{Architecture of co-attention encoder from \cite{xiong2016dynamic}.}
\label{coattention}
\end{figure}

\paragraph{Coattention} is proposed in \cite{xiong2016dynamic}. The architecture of the coattention encoder in DCN is shown in Fig.\ref{coattention}.
\par In the Coattention encoder, the affinity matrix $L = D ^ { \top } Q \in \mathbb { R } ^ { ( m + 1 ) \times ( n + 1 ) }$ is calculated and normalized row-wise and column-wise to obtain $A^Q$, the attention weights matrix across the document for each word of query, and $A^D$, the attention weights matrix across the query for each word of document. Then the attention contexts for question are computed $C ^ { Q } = D A ^ { Q } \in \mathbb { R } ^ { \ell \times ( n + 1 ) }$ and concatenated with $Q$ to obtain the final document representation $C ^ { D } = \left[ Q ; C ^ { Q } \right] A ^ { D } \in \mathbb { R } ^ { 2 \ell \times ( m + 1 ) }$. At the last step, [$D$,$C^D$] is fed to a bidirectional LSTM:
$$
u _ { t } = \mathrm { Bi-LSTM } \left( u _ { t - 1 } , u _ { t + 1 } , \left[ d _ { t } ; c _ { t } ^ { D } \right] \right) \in \mathbb { R } ^ { 2 \ell }
$$
The result serves as the foundation for predicting the answer. The hidden states form coattention encoding matrix $U =\left[ u _ { 1 } , \dots , u _ { m } \right] \in \mathbb { R } ^ { 2 \ell \times m }$. 
\par Similarly to Bi-directional Attention, the coattention mechanism utilizes attention information in two directions, while in a different way. It successively computes the attention contexts for the question and the document, and fuses them to get a co-dependent representation of document.

\paragraph{Self-matching Attention} is proposed in R-NET introduced before. Because many useful information exist in the passage context while they can not be captured by the traditional LSTM(which mainly exploits information in words' surrounding window), so the self-matching attention mechanism is proposed to address this problem. It collects evidence for each token $v_t$ from the whole passage and its according question information $c_t$ . And the result $h^P$ is the final passage representation:
$$
\begin{array} { l }
{h _ { t } ^ { P } = \operatorname { BiRNN } \left( h _ { t - 1 } ^ { P } , \left[ v _ { t } ^ { P } , c _ { t } \right]^* \right)} \\  
{\left[ v _ { t } ^ { P } , c _ { t } \right] ^ { * } = g _ { t } \odot \left[ v _ { t } ^ { P } , c _ { t } \right]}\end{array}
$$
$c_t$ here refers to an attention-pooling vector of the whole passage:
$$
\begin{array} { l } { s _ { j } ^ { t } = \mathrm { v } ^ { \mathrm { T } } \tanh \left( W _ { v } ^ { P } v _ { j } ^ { P } + W _ { v } ^ { \tilde { P } } v _ { t } ^ { P } \right) } \\ { a _ { i } ^ { t } = \exp \left( s _ { i } ^ { t } \right) / \Sigma _ { j = 1 } ^ { n } \exp \left( s _ { j } ^ { t } \right) } \\ { c _ { t } = \Sigma _ { i = 1 } ^ { n } a _ { i } ^ { t } v _ { i } ^ { P } } \end{array}
$$
and $g _ { t }$ is the gate define in Sec.\ref{ref_rnet}.
\par Uniquely, Self-matching Attention captures long-distance information from the passage itself. This helps R-NET in dealing with problems like coreference.

\subsection{Pre-trained word representations}
\label{sec:pretrain}
\par How to efficiently represent words as vectors, which serve as the base of most of the modern MRC systems, is a problem that concerns researchers very much. Previously, one-hot representation and N-gram model were popular, however, those simple techniques met their limits in many tasks. To address this problem, many technologies have been proposed. According to the time of occurrence, we introduce them as follows.

\begin{figure}
\centering
\includegraphics[width=1\textwidth]{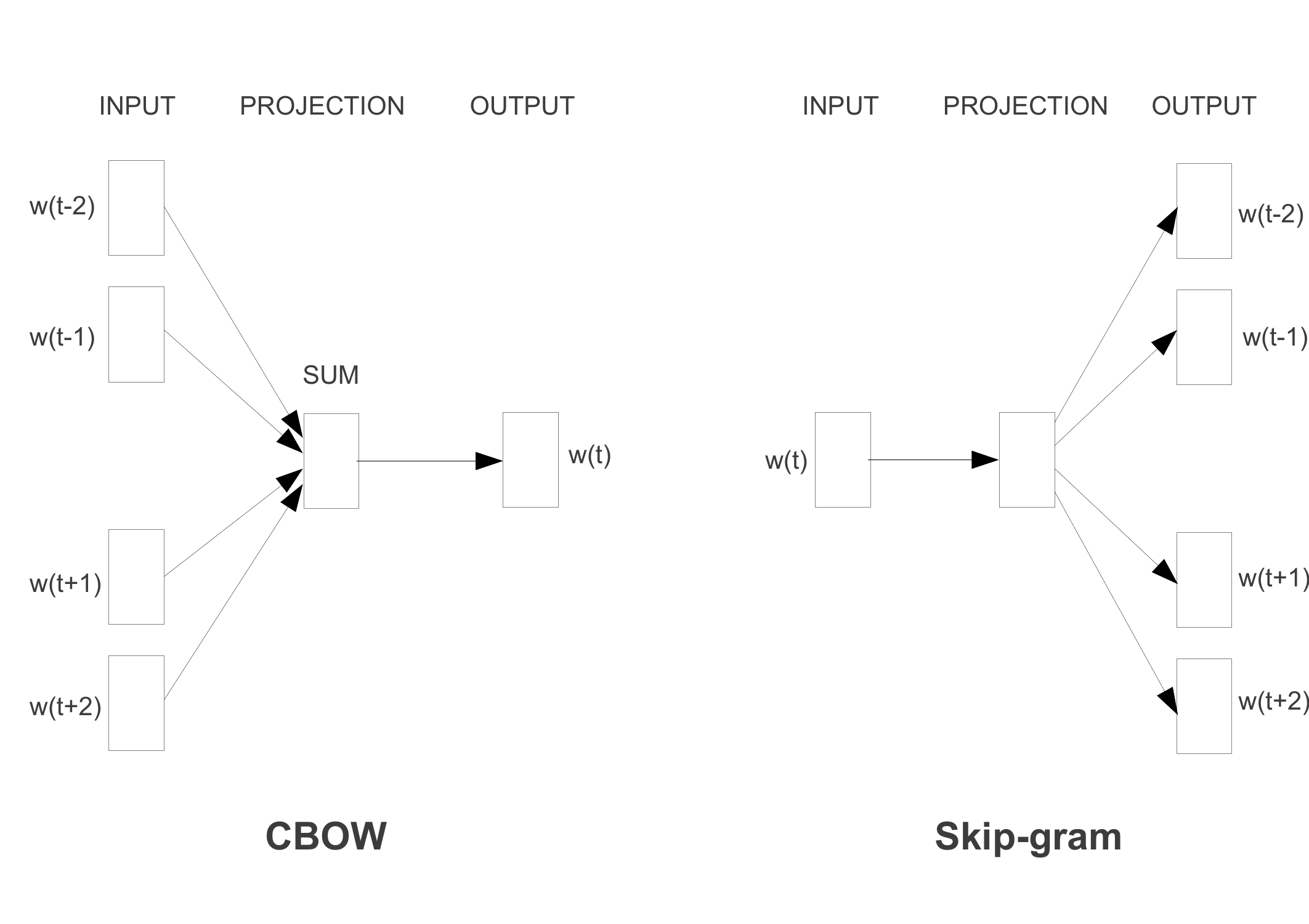}
\caption{Architectures of CBOW model and Skip-gram model from \cite{mikolov2013efficient}.}
\label{word2vec_pic_1}
\end{figure}

\paragraph{word2vec} Moving further from feedforward neural net language model(NNLM)\cite{bengio2003neural} and recurrent neural net language model(RNNLM), this paper\cite{mikolov2013efficient} proposed two novel models to learn the distributed representations of words, namely the Continuous Bag-of-Words Model(CBOW) and the Continuous Skip-gram Model. The architectures of these two models are given in Fig.\ref{word2vec_pic_1}.
\par The CBOW model uses several history words and future words as input and maximizes the probability of correctly predicting the current word. By contrast the skip-gram model uses current word as input and tries to predict words within a certain range before and after the current word. The result word vectors of both models achieved state-of-the-art performance on several tests.

\paragraph{GloVe} The word2vec method belongs to local context window methods, those methods can capture fine-grained semantic and syntactic regularities of words efficiently. However, they can not exploit global statistical information like latent semantic analysis(LSA)\cite{deerwester1990indexing}, which belongs to global matrix factorization methods. GloVe\cite{pennington2014glove} combines the advantages of these two family of methods.
\par GloVe takes the co-occurrence probabilities of words into consideration, and use the ratio of probabilities to reflect the relations of different words. For example, if we denote the  probability that word j appear in the context of word j as $P_{ij}$, then the ratio $ { P _ { i k } } / { P _ { j k } } $ can tell the correlation between certain words. An example is given in Fig.\ref{glove_pic_1}. The GloVe model $F$ takes the below form according to above phenomenon. 
$$
F \left( w _ { i } , w _ { j } , \tilde { w } _ { k } \right) = \frac { P _ { i k } } { P _ { j k } }
$$
where $w \in \mathbb { R } ^ { d }$ are word vectors. $F$ varies according to different constrains.

\begin{figure}
\centering
$$
\begin{tabular}{  l | l  l  l  l }
	 \text{Probability and Ratio} & k = solid & k = gas & k = water & k = fashion \\ \hline
	 P(k$|$ice) & $1.9 \times 10^{-4}$ & $6.6 \times 10^{-5}$ & $3.0 \times 10^{-3}$ & $1.7 \times 10^{-5}$  \\ 
	 P(k$|$steam) & $2.2 \times 10^{-5}$ & $7.8 \times 10^{-4}$ & $2.2 \times 10^{-3}$ & $1.8 \times 10^{-5}$   \\ 
	 P(k$|$ice)/P(k$|$steam) & 8.9 & $8.5 \times 10^{-2}$ & 1.36 & 0.96 \\ 
\end{tabular}
$$
\caption{from \cite{pennington2014glove}. A ratio much greater than 1 means word k correlate well with ice, and a ratio much greater than 1 means word k correlate well with stream.}
\label{glove_pic_1}
\end{figure}

\paragraph{ELMo} One disadvantages of word vectors generated by above methods is that they are static, thus are independent of application linguistic contexts. This may lead to poor performance when it comes to polysemy. In light of this, ELMo\cite{peters2018deep} was proposed to addresses this problem. 
\par ELMo's model employs a bi-LSTM\cite{hochreiter1997long} with character convolutions on the input.
$$
\begin{aligned}
p \left( t _ { 1 } , t _ { 2 } , \ldots , t _ { N } \right) &= \prod _ { k = 1 } ^ { N } p \left( t _ { k } | t _ { 1 } , t _ { 2 } , \ldots , t _ { k - 1 } \right) \\  
p \left( t _ { 1 } , t _ { 2 } , \ldots , t _ { N } \right) &= \prod _ { k = 1 } ^ { N } p \left( t _ { k } | t _ { k + 1 } , t _ { k + 2 } , \ldots , t _ { N } \right)
\end{aligned}
$$
Then it jointly maximizes the log likelihood of the forward and backward directions and record the internal states. 
$$
\begin{aligned}   \sum _ { k = 1 } ^ { N } ( &\log p ( t _ { k } | t _ { 1 } , \ldots , t _ { k - 1 } ; \Theta _ { x } , \stackrel { \rightarrow } { \Theta } _ { L S T M } , \Theta _ { s } )   \\  &+ \log p ( t _ { k } | t _ { k + 1 } , \ldots , t _ { N } ; \Theta _ { x } ,  \stackrel { \leftarrow } { \Theta } _ { L S T M } , \Theta _ { s } ) )  \end{aligned}
$$

Finally a task specific linear combination of those internal states are used to obtain the ELMo representation. In this way, ELMo can capture context-dependent aspects of word meaning as well as syntax information for each token. If fine-tuned on domain specific data, the model usually performs better.

\begin{figure}
\centering
\includegraphics[width=1\textwidth]{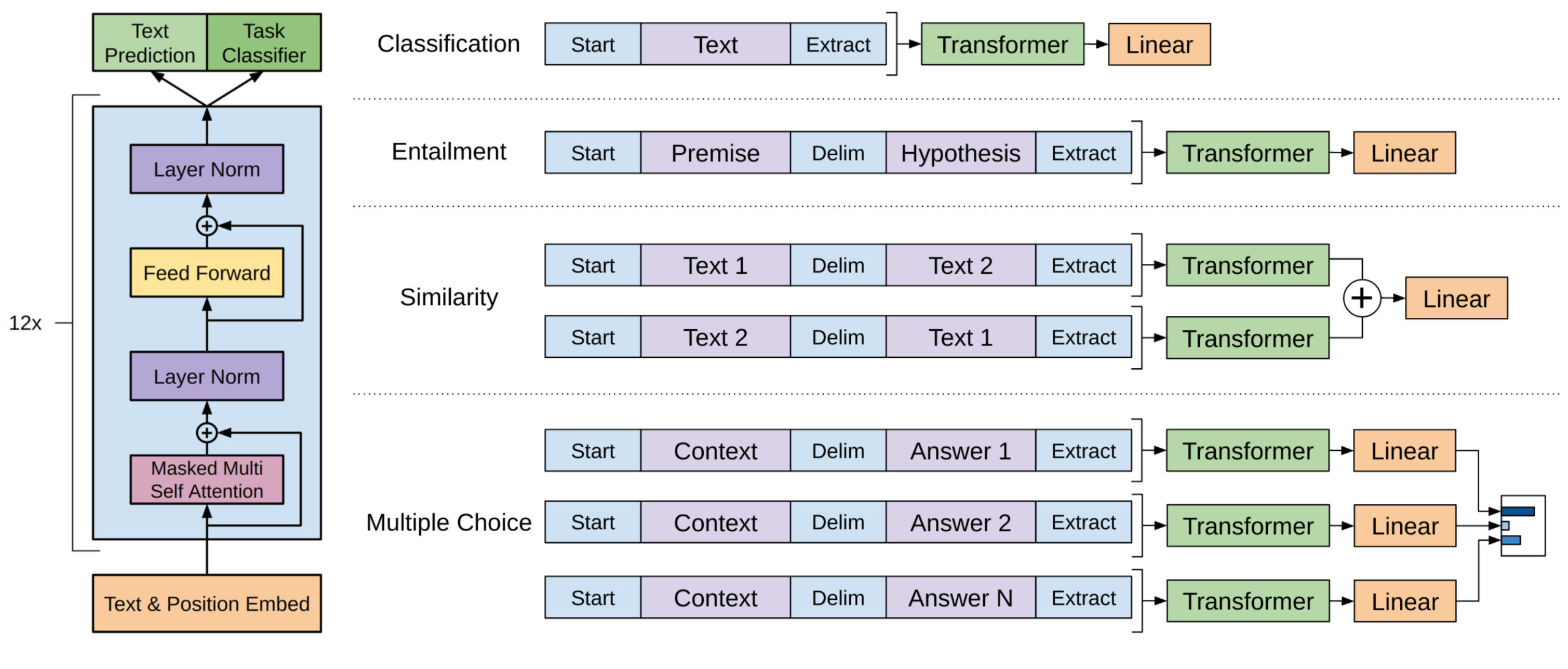}
\caption{Graph comes from paper\cite{radford2018improving}. Left is transformer architecture and training objectives used in this work. Right is input transformations for fine-tuning on different tasks. All structured inputs are converted into token sequences to be processed by GPT, followed by a linear+softmax layer.
}
\label{gpt_pic_1}
\end{figure}

\paragraph{GPT} Compared to ELMo, GPT\cite{radford2018improving} uses a variant of Transformer\cite{vaswani2017attention} instead of LSTM to better capture the long term linguistic structure. The overview of this work is given in Fig.\ref{gpt_pic_1}. Given a corpus $\mathcal { U } = \left\{ u _ { 1 } , \dots , u _ { n } \right\}$, a standard language model with a multi-layer Transformer decoder\cite{liu2018generating} is used:
$$
L _ { 1 } ( \mathcal { U } ) = \sum _ { i } \log P \left( u _ { i } | u _ { i - k } , \ldots , u _ { i - 1 } ; \Theta \right)
$$
$$
\begin{aligned} h _ { 0 } & = U W _ { e } + W _ { p } \\ 
h _ { l } & = \operatorname { transformer\_block } \left( h _ { l - 1 } \right) \forall i \in [ 1 , n ] \\ 
P ( u ) & = \operatorname { softmax } \left( h _ { n } W _ { e } ^ { T } \right) \end{aligned}
$$
where $k$ is the context window size, $U = \left( u _ { - k } , \ldots , u _ { - 1 } \right)$ is the context vectors of tokens, $n$ is the number of layers, $W_e$ is the token
embedding matrix, and $W_p$ is the position embedding matrix. All the parameters are trained using stochastic gradient descent\cite{robbins1985stochastic}. The final transformer block’s activation is denoted as $h^m_l$.
\par A supervised fine-tuning can be applied in different down-stream tasks. As for some tasks like text classification, only a linear output layer with parameters $W_y$ is needed to predict $y$:
$$
P ( y | x ^ { 1 } , \ldots , x ^ { m } ) = \operatorname { softmax } \left( h _ { l } ^ { m } W _ { y } \right)
$$
\par More recently, its successor GPT2 is released, which is a scale-up of GPT while with much larger volume. GPT2 has 1.5 billion parameters, and claimed to achieve state-of-the-art performance on many language modeling. However its code have not been released by the time this paper is written.

\begin{figure}
\centering
\includegraphics[width=1\textwidth]{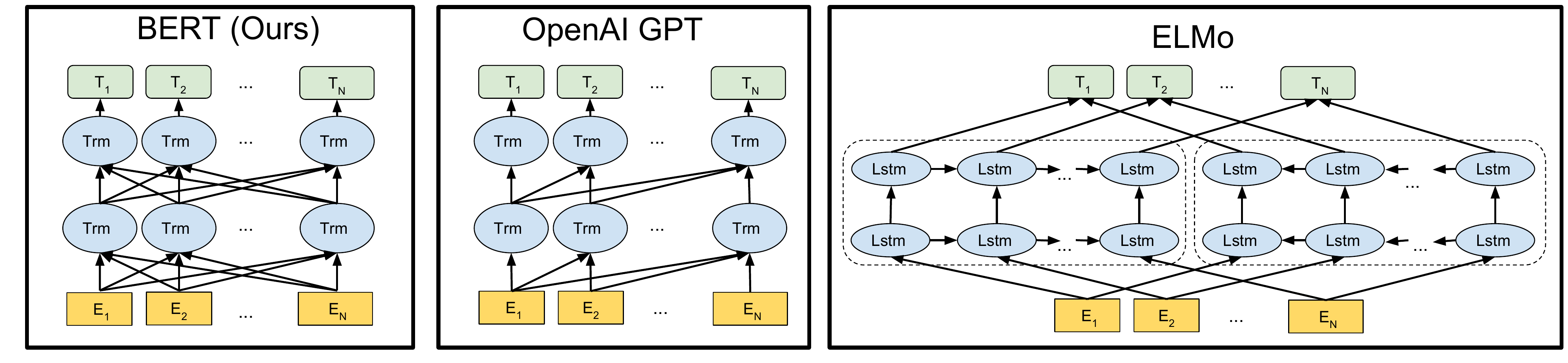}
\caption{Model Architectures of BERT, GPT and ELMo  Quoted from \cite{devlin2018bert}}
\label{bert_pic_1}
\end{figure}

\paragraph{BERT} As shown in Fig.\ref{bert_pic_1}, both ELMo and GPT models only use unidirectional language models to learn the representation of tokens. BERT\cite{devlin2018bert} points out that this restriction has severely limited the efficiency of the pre-trained representation. To address this problem, two new prediction tasks are proposed to pre-train BERT in two direction, namely the "masked language model" and the "Next Sentence Prediction".
\par Inspired by the Cloze\cite{ref_cloze} task, the "masked language model" is to predict the randomly masked tokens' id based on their context in the input. In other words, both the left and the right context will be taken into consideration when computing representations. And to capture sentence level information and relationship, a binarized "Next Sentence Prediction" task is to predict whether a sentence $A$ is the next sentence of $B$.
\par
\begin{figure}
\centering
\includegraphics[width=1\textwidth]{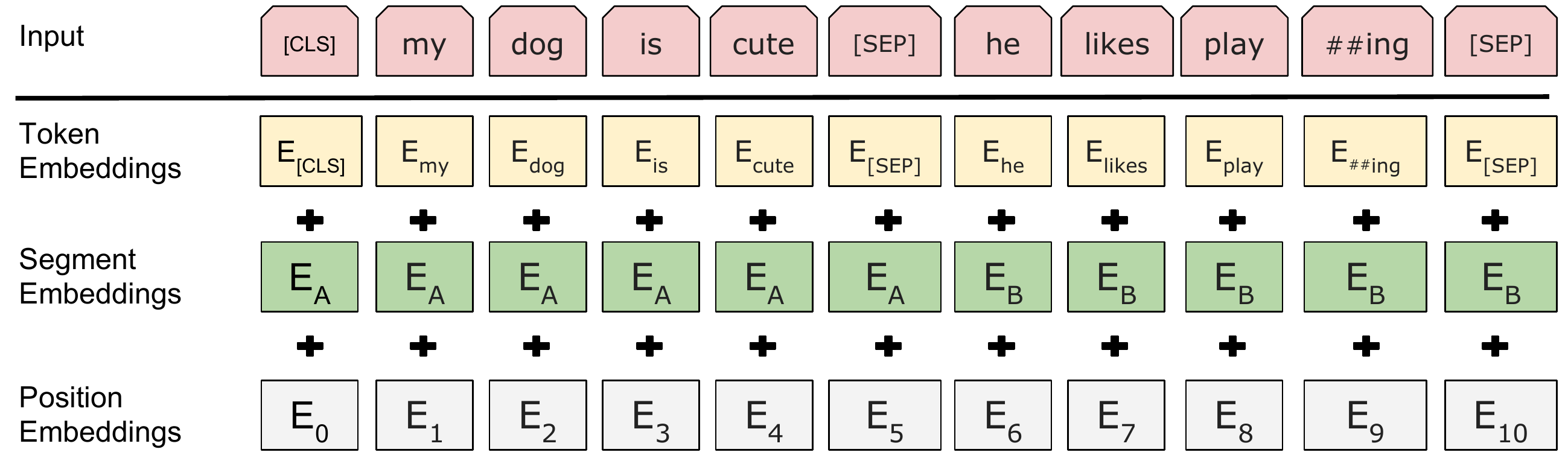}
\caption{BERT Input Representation \cite{devlin2018bert}.}
\label{bert_pic_2}
\end{figure}

 The WordPiece embeddings\cite{wu2016google} are used in the input layer along with the Segment Embeddings and the Position Embeddings. The input embeddings is the sum of above three embeddings, as shown in Fig.\ref{bert_pic_2}. The main architecture of BERT's model is a multi-layer bidirectional Transformer encoder almost identical to the original one\cite{vaswani2017attention}.
\par
\begin{figure}
\centering
\includegraphics[width=1\textwidth]{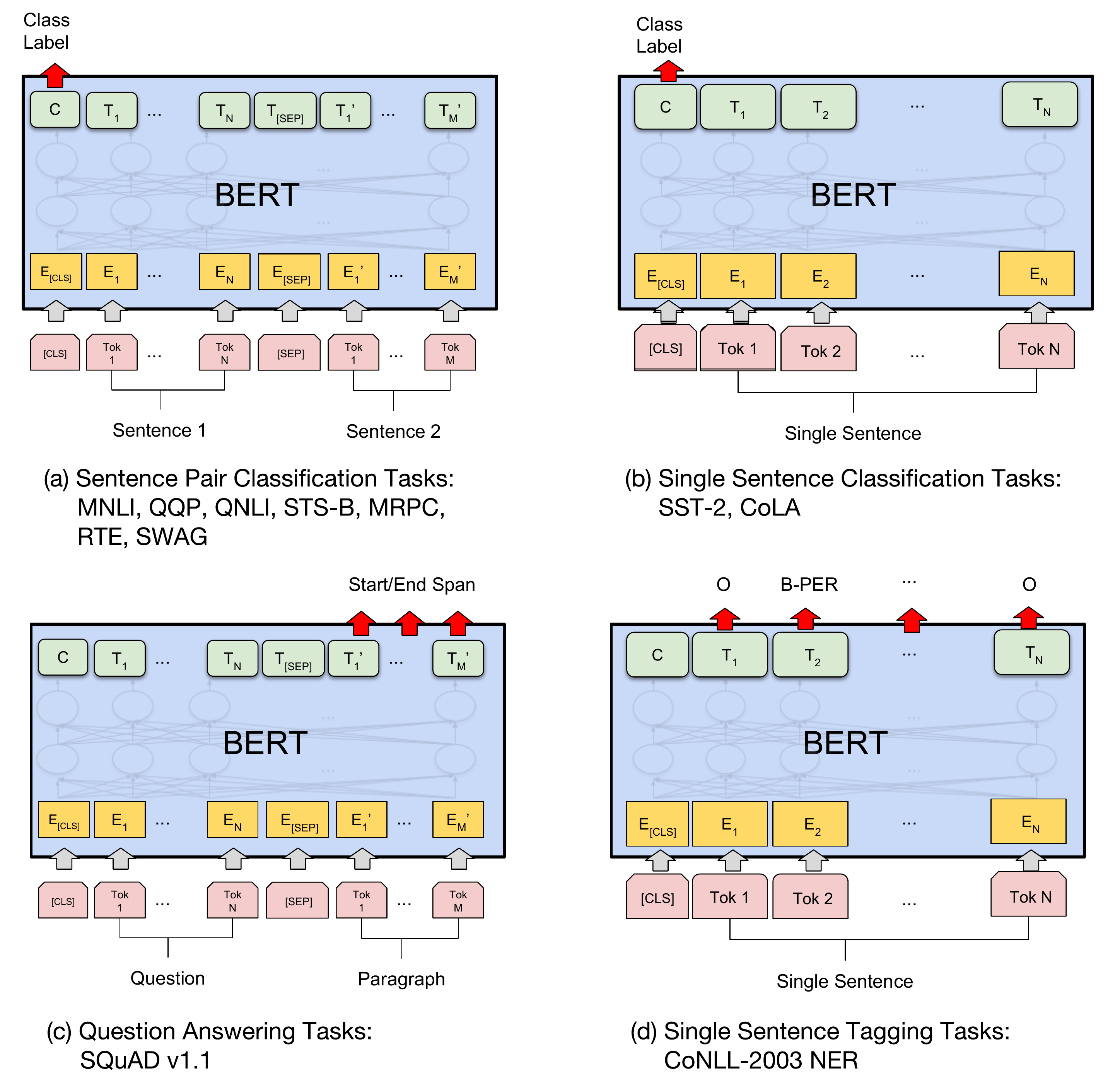}
\caption{Task specific models overview from paper\cite{devlin2018bert}.}
\label{bert_pic_3}
\end{figure}

 Similar to GPT, when fine-tuned on down-steam tasks, only an additional output layer with a minimal number of parameters is needed, as shown in Fig.\ref{bert_pic_3}. BERT advanced state-of-the-art results on 11 NLP tasks.
\par A comparison of size of BERT and GPT is given in Table \ref{compare_4}.

\begin{table}
    \centering
    \setlength{\tabcolsep}{8mm}{
    \begin{tabular}{lccc}
    \toprule
         Model & Parameters & Layers & Hidden size\\
         \midrule 
         $GPT$ &  117M & 12 & 768\\
         $BERT_{BASE}$ & 110M & 12 & 768 \\
         $BERT_{LARGE}$ & 340M & 24 & 1024 \\
         $GPT2$ & 1542M & 48 & 1600\\
         \bottomrule
    \end{tabular}}
    \caption{Hyperparameter Comparison among 4 Similar Models. Layers means the number of the transformer blocks.}
    \label{compare_4}
\end{table}

\section{Conclusion}
In this paper, we summarized advances in MRC field in recent years. In section\ref{intro}, we briefly introduced the history of MRC tasks and some early MRC systems. In section \ref{sec:corpus}, we introduced recent datasets in three categories, i.e. SQuAD, CNN/Daily mail, CBT, NewsQA, TriviaQA
and CLOTH in Extractive format, MS MARCO and Narrative QA in Narrative format and WIKIHOP, MCTest, RACE, MCScript and ARC in Multiple-choice format. The CoQA, a novel dataset focuses on conversational questions is also included.
\par In section \ref{sec:3}, we first go through several non-neural methods, including Sliding Window, Logistic regression, TF-IDF and Boosted method, then more importantly the neural-based models like mLSTM+Ptr, DCN, GA, BiDAF, FastQA, RNET, ReasoNet and QAnet. Afterwards we discussed and compared two important compositions of these models, namely the Pre-training technology and Attention mechanism, in detail. We covered Word2Vec, Glove, ELMo, GPT\&GPT2 and BERT in section \ref{sec:pretrain}, and hard attention, soft attention, Bi-directional attention, coattention and self-attention mechanisms in section \ref{sec_attention}.
\par All together, we reviewed the major progress that has been made in recent years in MRC field. However, the MRC direction is developing very fast and it is difficult to include all the newly proposed MRC work in this survey. We hope this review will ease the reference to recent MRC advences, and encourage more researchers to work on MRC field.

\bibliographystyle{spmpsci}      % mathematics and physical sciences
\bibliography{thesis}   

\end{document}